\documentclass{article}

\usepackage[preprint]{neurips_2026}
\usepackage[utf8]{inputenc} % allow utf-8 input
\usepackage[T1]{fontenc}    % use 8-bit T1 fonts
\usepackage{hyperref}       % hyperlinks
\usepackage{url}            % simple URL typesetting
\usepackage{booktabs}       % professional-quality tables
\usepackage{amsfonts}       % blackboard math symbols
\usepackage{nicefrac}       % compact symbols for 1/2, etc.
\usepackage{microtype}      % microtypography
\usepackage[table]{xcolor}   % colors
\usepackage{graphicx}
\usepackage{subcaption}
\usepackage{caption}
\usepackage{capt-of}
\usepackage[textsize=tiny]{todonotes}
\usepackage[most]{tcolorbox}
\usepackage{enumitem}
\usepackage{xspace,soul}
\usepackage{duckuments}
\usepackage{lipsum}
\usepackage{multirow}
\definecolor{xblue}{HTML}{4169E1}
\definecolor{xgreen}{HTML}{036C3A}
\definecolor{xpurple}{HTML}{9838B1}
\definecolor{xslategray}{HTML}{70818F}
\definecolor{xorange}{HTML}{FF8C00}
\definecolor{xcyan}{HTML}{06AEEF}
\definecolor{xred}{HTML}{FF0000}
\definecolor{xgray}{HTML}{808080}
\definecolor{xxgreen}{HTML}{009F86}
\definecolor{xsienna}{HTML}{8B4512}
\definecolor{xxpurple}{HTML}{623E99}

\newcommand{\xblue}[1]{\textcolor{xblue}{#1}}

\newcommand{\xslategray}[1]{\textcolor{xslategray}{#1}}
\newcommand{\xxpurple}[1]{\textcolor{xxpurple}{#1}}

\newcommand{\coloredhl}[2]{{\textcolor{#1}{\sethlcolor{#1!10}\hl{#2}}}\xspace}
\newcommand{\coloredul}[2]{\textcolor{#1}{\underline{#2}}\xspace}

\def\rvx{{\mathbf{x}}}

\def\rvz{{\mathbf{z}}}

\newcommand{\mask}{\mathbf{m}}
\usepackage{algorithm}
\usepackage[noend]{algpseudocode}
\newcommand{\algindent}{\hspace{\algorithmicindent}}

\algrenewcommand{\algorithmiccomment}[1]{\hfill \xslategray{\texttt{\#~#1}}}
\usepackage[most]{tcolorbox}
\usepackage{wrapfig}

\usepackage{fvextra}
\usepackage{xcolor}
\usepackage{tabularx}

\newenvironment{promptbox}
{\VerbatimEnvironment
\begin{Verbatim}[fontsize=\scriptsize, frame=single, framesep=4pt,
breaklines, breakanywhere]}
{\end{Verbatim}}

\title{Blackboard Intelligence Can Surpass Autoregressive on Globally Constrained Problems}
\author{
Woosang Jeon$^{1}$\thanks{Equal contribution; lead junior authors. $^\dagger$ Lead senior authors.} \quad
Jaeyeon Kim$^{2}$\footnotemark[1] \quad
Sham Kakade$^{2}$ \quad
Yilun Du$^{2}$ \quad
Amrit Singh Bedi$^{3}$
\AND
Arun Kumar Chithanar \quad
Chul Lee \quad
Taehyeong Kim$^{1}$\footnotemark[2]\quad
Sitan Chen$^{2}$\footnotemark[2]
\\[1.7em]
{
$^{1}$Seoul National University
\quad
$^{2}$Harvard University
\quad
$^{3}$University of Central Florida
}
}
\begin{document}
\maketitle
\vspace{-0.1in}
\begin{abstract}
Next-token prediction has driven remarkable progress in large language models, yet a growing body of evidence suggests that they can struggle on problems governed by complex global constraints.
In this work, we focus on this regime and ask whether some of these limitations arise from the inference interface induced by next-token prediction itself.
We study this question through \emph{blackboard intelligence}: an inference-time perspective in which a model works on a fixed, revisable canvas and searches over candidate solution states rather than committing to a causal, left-to-right trajectory.
We instantiate this idea with diffusion language models, whose any-order prediction interface naturally exposes predictions over partially filled solution states.
Our key observation is that \emph{mean confidence}, a simple model-internal quantity available from the standard masked diffusion objective, provides a useful proxy for global coherence and can guide inference-time search and revision.
Empirically, across ZebraLogic, Nurse Rostering, and Job-Shop Scheduling, Blackboard consistently improves inference while holding the fine-tuned LLaDA-8B-Instruct checkpoint fixed and substantially outperforms same-scale autoregressive baselines, reaching $90.4\%$ accuracy on ZebraLogic-Hard, $76.4\%$ exact feasibility on Nurse Rostering, and $80.2\%$ optimality on JSSP.
Stronger autoregressive search and refinement also fail to close the gap on ZebraLogic-Hard, while Blackboard surpasses tested frontier LLMs there and on JSSP despite their substantially greater scale and strong test-time reasoning.  We open-source our codebase at \url{https://github.com/jwoosang1/blackboard-intelligence}.
\end{abstract}
% {left bottom right top}
\begin{figure}[h]
    \centering
    \vspace{-0.10in}
\includegraphics[width=0.9\linewidth,trim={0.2cm 0.2cm 0.2cm 0.0cm}, clip]{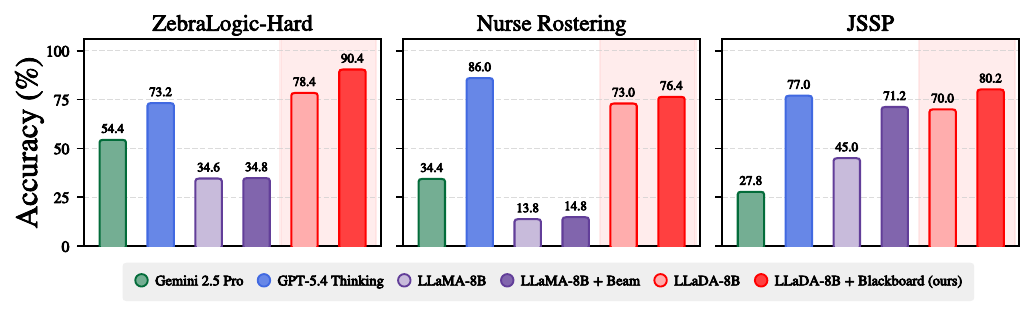}
\caption{
Across three globally constrained domains, \emph{blackboard} inference consistently improves LLaDA-8B-Instruct over its standard inference (shaded) and outperforms same-scale autoregressive baselines; on ZebraLogic-Hard and JSSP, it also exceeds the tested frontier LLMs.
}
    \label{fig:main_result}
    \vspace{-0.20in}
\end{figure}

\section{Introduction}
\label{sec:intro}

In recent years, autoregressive next-token prediction has been the dominant modeling paradigm underlying Large Language Models (LLMs).
This paradigm has delivered remarkable gains in language modeling, coding, mathematics, and agentic tasks~\citep{achiam2023gpt,jaech2024openai,comanici2025gemini,guo2025deepseek,anthropic2025}, leading to the view that sufficiently advanced LLMs can solve arbitrarily complex problems. 
At the same time, a growing body of evidence suggests that this perspective may be incomplete; LLMs can struggle in domains governed by complex combinatorial or logical constraints~\citep{nagarajan2025roll,lin2025zebralogic,abgaryan2025starjob,tso2026constraintbench,fesser2026evaluating}.
 
In this work, we ask whether some of these limits arise not merely from model scale or data, but also from the inference interface induced by next-token prediction. We argue that models that instead operate on a fixed canvas and natively support any-order unmasking and remasking to search over the solution space can offer an alternative inference interface in domains where next-token prediction struggles.
We informally refer to this canvas-based, revision-capable inference as \xblue{\textbf{\emph{blackboard intelligence}}}: the model iteratively builds and revises a candidate solution, resembling how humans solve structured problems by dynamically maintaining and searching over partial solutions.

Diffusion language models, specifically masked diffusion models~\citep {shi2024simplified,sahoo2024simple,nie2025large}, offer a natural instantiation of blackboard intelligence. While recent large-scale diffusion language models have been studied primarily through the lens of inference-time efficiency~\citep{gemini2025diffusion,wu2025fast2,wu2025fast,labs2025mercury,song2025seed}, in this work, we shift the focus from \emph{speed to intelligence}: we ask whether their flexible inference interface can support \emph{problem-solving capabilities} difficult to realize with standard autoregressive generation.

\textbf{Contribution.} 
Our central claim is that the standard masked diffusion objective gives rise to a state-level inference interface that enables diffusion language models to exhibit blackboard intelligence on globally constrained problems.
In particular, we show that a simple notion of \emph{mean confidence} provides a useful model-internal signal of the global coherence of the current state.
Unlike next-token prediction, masked diffusion makes predictions over arbitrary unrevealed, i.e., masked, positions, making such a state-level signal natively available.
We use this mean confidence to guide inference-time search, revision, and adaptive intervention over a fixed canvas.

Empirically, without changing the fine-tuned LLaDA-8B-Instruct~\citep{nie2025large} checkpoint, Blackboard improves standard inference across three domains that involve global constraints: ZebraLogic, Nurse Rostering, and JSSP, reaching $90.4\%$ accuracy on ZebraLogic-Hard, $76.4\%$ exact feasibility on Nurse Rostering, and $80.2\%$ optimality on JSSP (Figure~\ref{fig:main_result}).
Next, we show that stronger autoregressive inference does not recover comparable gains in our evaluation, even with substantially increased inference-time computation.
On ZebraLogic-Hard, LLaMA-3.1-8B Instruct~\citep{meta2024llama} reaches only $35.0\%$ with wider beam search and $36.6\%$ with verifier-guided refinement, while frontier autoregressive models such as GPT-5.4~\citep{openai2026} and Gemini 2.5 Pro~\citep{google2025} also remain below Blackboard on ZebraLogic-Hard and JSSP despite extensive inference-time reasoning.
% Together, these results support the hypothesis that direct access to a persistent, revisable partial solution state provides an important inference-level advantage on globally constrained problems.

\textbf{Implication.}
Our empirical results point to an interface-level distinction between autoregressive and masked-diffusion-based generation: the masked-diffusion interface can enable a distinct form of problem solving based on \emph{blackboard intelligence}, in which a persistent partial solution can be evaluated and revised throughout inference.
This form of state-level inference control is not directly exposed by a standard next-token decoding pass, and the autoregressive search and refinement procedures do not recover comparable gains from additional inference-time computation.

\textbf{Organization.}
Section~\ref{sec:prelim} reviews autoregressive and diffusion LLMs.
Section~\ref{sec:challenge} introduces mean confidence and empirically demonstrates its effectiveness as a state-level reliability signal.
Building on this signal, Section~\ref{sec:experiments} develops an algorithmic instantiation of \emph{blackboard intelligence} and evaluates it across the three domains.
\vspace{-0.10in}
\section{Preliminaries and Framework}
\label{sec:prelim}
\vspace{-0.10in}
In this section, we briefly review the training and inference procedures of both autoregressive and diffusion language models. Let $p_{\mathrm{data}}$ denote the target distribution over discrete sequences that a generative model $f_\theta$ aims to sample from, with vocabulary $\mathcal{V}$ and sequence length $L$. For a sequence $\rvx=(\rvx^1,\dots,\rvx^L)\in\mathcal{V}^L$, we write $\rvx^{<i}:=(\rvx^1,\dots,\rvx^{i-1})$ for its prefix before position $i$.

\textbf{Autoregressive LLMs.}
Autoregressive LLMs factorize the data distribution causally as $p_{\mathrm{data}}(\rvx)=\prod_{i=1}^L p_{\mathrm{data}}(\rvx^i \,|\,\rvx^{<i})$. Accordingly, an autoregressive model $f_\theta$ is trained to approximate the next-token posterior $f_\theta(\cdot\,|\,\rvx^{<i})\approx p_{\mathrm{data}}(\rvx^i \,|\,\rvx^{<i})$, typically by minimizing the cross-entropy loss.
At inference time, starting from a prompt $\hat{\rvx}^{<1}=[\texttt{prompt}]$ (or empty prefix), we iteratively sample $v\sim f_\theta(\cdot \mid \hat{\rvx}^{<i})$ and append it to form $\hat{\rvx}^{<i+1}$.

\textbf{Diffusion LLMs: training.}
Diffusion LLMs (dLLMs), instantiated by masked diffusion models~\citep{sahoo2024simple,shi2024simplified,lou2023discrete}, learn to predict the clean-token posterior at \emph{all unrevealed positions} from partially observed sequences, rather than predicting only the next token. 

Formally, let $\rvx\sim p_{\mathrm{data}}$ be a clean sequence. We sample an integer $n\sim\mathrm{Unif}\{1,\dots,L\}$, choose a subset $M\subseteq\{1,\dots,L\}$ of size $n$ uniformly at random, and construct a masked sequence $\rvz\in(\mathcal{V}\cup\{\mask\})^L$ by replacing the indices in $M$ with an auxiliary mask token $\mask$. This masking procedure induces a joint distribution over $(\rvx,\rvz)$ and, for each masked position $i\in M$, the associated coordinate-wise posterior is $p(\rvx^i=v\,|\,\rvz)$. The masked diffusion model, parameterized by $f_\theta$, takes $\rvz$ as input and outputs a categorical distribution $f_\theta^i(\cdot\,|\,\rvz)\approx p(\rvx^i=\cdot\,|\,\rvz)$ for each position $i$. The training objective is the cross-entropy loss over the masked positions.

\vspace{-0.06in}
\begin{equation*}
    \mathcal{L}(\theta)
    \;=\;
    \mathbb{E}_{\rvx,\rvz}
    \left[\frac{1}{|M|}\sum_{i\colon \rvz^i=\mask}
        -\log f_\theta^i(\rvx^i \,|\,\rvz) \right],
\end{equation*}
\vspace{-0.06in}

\textbf{Diffusion LLMs: inference.}
dLLM inference starts from a length-$L$ masked sequence $\rvx_1 = (\mask,\dots,\mask)$ or generally with a given prompt $\rvx_1 = ([\texttt{prompt}],\mask,\dots,\mask)$. It proceeds over a monotonically decreasing time grid $t_0=1>\dots>t_N = 0$. At each step $t_\ell$, given a partially masked sequence $\rvx_{t_\ell}\in(\mathcal{V}\cup \{\mask\})^L$, we proceed in two steps to obtain $\rvx_{t_{\ell+1}}$: \textbf{(a)} Choose a subset of masked tokens $\mathcal{S}$ and \textbf{(b)} For each $i \in \mathcal{S}$, unmask $\rvx_{t_\ell}^i$ to a clean token sampled from $f_\theta^i(\cdot\,|\, \rvx_{t_\ell})$.

Notably, the choice of $\mathcal{S}$ in step (a) is highly flexible and central to dLLMs' gains on downstream tasks. Since the model produces a predictive distribution
$f_\theta^i(\cdot\,|\, \rvx_{t_\ell})\in\Delta(\mathcal{V})$
for every masked position $i$, it enables informed choices of $\mathcal{S}$ at each step.

A standard way to instantiate this flexibility is \emph{greedy decoding}, also referred to as \emph{confidence-based decoding}. This strategy is especially common in dLLMs, as it is easy to deploy and is practically effective on downstream benchmarks
~\citep{nie2025large,kim2025train,zheng2024masked,peng2025pathplanningmaskeddiffusion,ben2025accelerated,hayakawa2025demystifying}. Specifically, the sampler computes a confidence score for every masked position and selects the highest-scoring positions,
$\mathcal{S}=
\mathrm{TopK}_{i:\,\rvx_t^i=\mask}
[\mathrm{score}(i)]$,
where $\mathrm{score}(i)$ quantifies how certain the model is about its prediction at position $i$. Common choices include the maximum predicted probability
$\max_{v\in\mathcal{V}} f_\theta^i(v\,|\,\rvx_t)$,
the margin between the top two probabilities
$f_\theta^i(v_1\,|\,\rvx_t)-f_\theta^i(v_2\,|\,\rvx_t)$,
where $v_1$ and $v_2$ are the most and second-most likely tokens, and the negative entropy of the categorical distribution.

% \amrit{may be dLLM inference is the most important part for us, and explaining it with the help of a pseudo code might help the reader better?}

\section{Mean Confidence as a Probe of Global Coherence} 
\label{sec:challenge}

\subsection{Next-token prediction under global constraints}
\label{sec:llm_limit}

\paragraph{Curse of complexity in LLM reasoning.}

\begin{figure}[ht]
    \centering
\includegraphics[width=\linewidth,trim={0.2cm 0.2cm 0.2cm 0.2cm}]{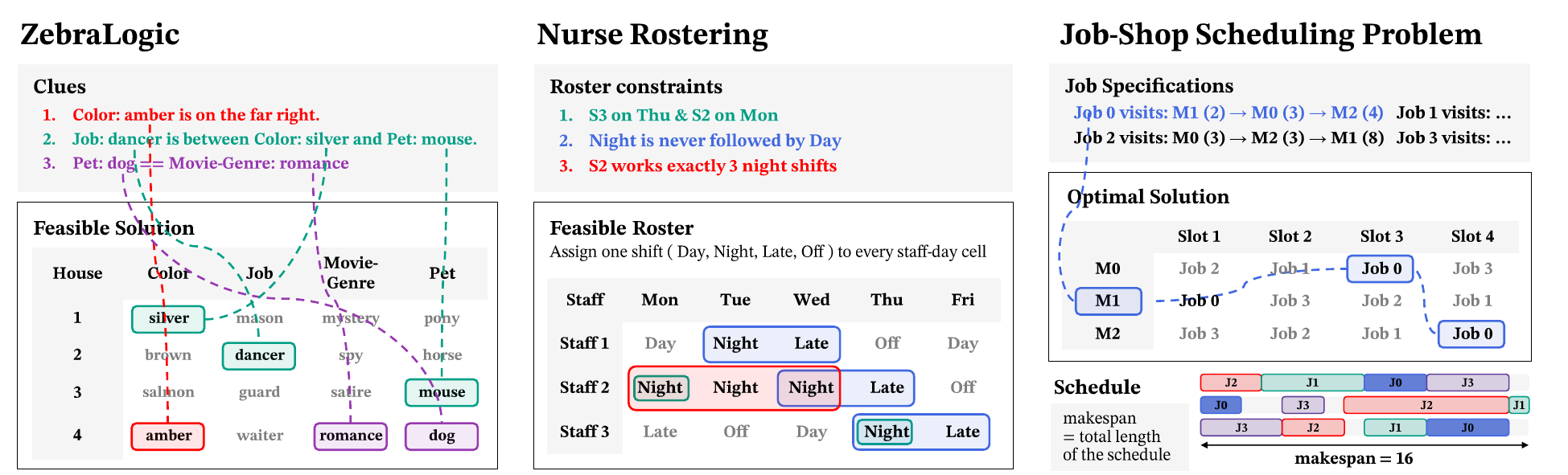}
\caption{\textbf{Illustration of globally constrained tasks}. ZebraLogic asks for a feasible assignment satisfying all clues; Nurse Rostering asks for a feasible staff--day shift assignment satisfying roster constraints; and JSSP asks for a feasible schedule with globally minimum makespan.}
    \label{fig:task_exp}
    % \vspace{-0.14in}
\end{figure}

\begin{wrapfigure}[14]{r}{0.47\textwidth}   
\centering
\includegraphics[width=0.45\textwidth,trim={0.5cm 0.5cm 0.5cm 0.5cm}]{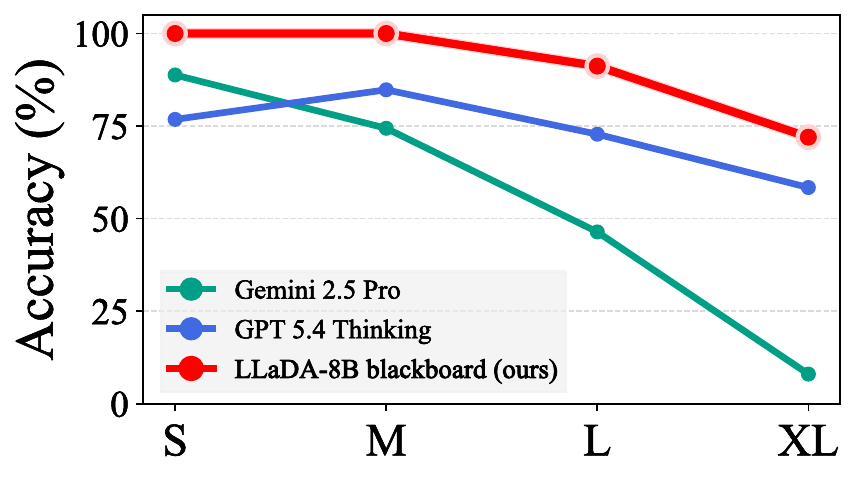}
\caption{\textbf{Curse of complexity on ZebraLogic}. Frontier LLM accuracy degrades sharply with difficulty; LLaDA-8B blackboard remains high.}
\label{fig:zl_curse}
\end{wrapfigure}

We first revisit the empirical finding of~\cite{lin2025zebralogic}. 
ZebraLogic is a family of logic-grid puzzles derived from constraint satisfaction problems. 
% Each instance specifies a collection of entities, attributes, and natural-language clues, and the task is to recover the unique global assignment that satisfies all clues simultaneously (Figure~\ref{fig:task_exp}, left). 
Each instance specifies entities, attributes, and natural-language clues, and the task is to recover the unique assignment satisfying all clues simultaneously (Figure~\ref{fig:task_exp}, left).
% A filled solution board is treated as correct only if all clues are satisfied, i.e., there is no partial credit. 
% As the problem difficulty scales from ZebraLogic-S to ZebraLogic-XL, the underlying search space grows exponentially, and the performance of state-of-the-art LLMs deteriorates sharply (Figure~\ref{fig:zl_curse}).
As difficulty scales from ZebraLogic-S to ZebraLogic-XL, the search space grows exponentially and state-of-the-art LLM performance deteriorates sharply (Figure~\ref{fig:zl_curse}).

We additionally consider Nurse Rostering, a constrained feasibility problem that retains ZebraLogic's requirement of satisfying all constraints while introducing an operational scheduling structure.
The task assigns staff to shifts across multiple days under coupled coverage, workload, and temporal constraints (Figure~\ref{fig:task_exp}, middle).
A similar difficulty trend emerges as conflicts among these constraints become denser (Appendix~\ref{app:nr_by_difficulty}).

% In this work, we also consider an even more challenging combinatorial optimization problem: the Job-Shop Scheduling Problem (JSSP). In JSSP, each job consists of an ordered sequence of operations, where each operation must be processed on a specified machine for a specified duration. 
% The goal is to construct a feasible schedule satisfying all precedence and machine-capacity constraints, while minimizing the \emph{makespan}, defined as the completion time of the last operation, or equivalently, the total length of the schedule. 
% A completed schedule is considered \emph{globally optimal} only if it attains the minimum makespan among all feasible schedules (Figure~\ref{fig:task_exp}, right). 
% Unlike ZebraLogic, JSSP requires not only finding a feasible assignment \emph{but also} searching for the global optimum among those. 
% We observe a similar degradation in LLM performance as JSSP complexity increases (Appendix~\ref{app:jssp_by_size}).
Finally, we consider the Job-Shop Scheduling Problem (JSSP), which extends this scheduling structure from constrained feasibility to combinatorial optimization.
In JSSP, each job consists of an ordered sequence of operations, where each operation must be processed on a specified machine for a specified duration.
The goal is to construct a feasible schedule satisfying all precedence and machine-capacity constraints while minimizing the \emph{makespan}, the completion time of the last operation.
Unlike ZebraLogic and nurse rostering, JSSP requires not only finding a feasible assignment but also searching for a globally optimal one (Figure~\ref{fig:task_exp}, right).
The difficulty trend remains pronounced under this optimization objective, particularly as problem size grows (Appendix~\ref{app:jssp_by_size}).

%

% {left bottom right top}

\textbf{Limit of next-token prediction.}
% We view this \emph{curse of complexity} as evidence of a structural limitation of next-token prediction, especially on reasoning problems that require global structure. 
We view this \emph{curse of complexity} as consistent with an interface-level limitation of next-token prediction, especially on problems that require global structure.
% ZebraLogic and JSSP \emph{can be efficiently approached by} maintaining a partially filled answer, backtracking when constraints are violated, and branching over multiple candidates. 
Such globally constrained problems \emph{can be efficiently approached by} maintaining a partially filled answer, backtracking when constraints are violated, and branching over multiple candidates.
% Autoregressive models, however, operate through a serialized trace, making such procedures cumbersome: changing an earlier decision requires recovering information from a long history and regenerating what follows. 
Autoregressive prediction, however, operates through a serialized trace, making such procedures cumbersome: changing an earlier decision requires recovering information from a long history and regenerating what follows.
Moreover, because the next-token prediction objective supervises only \emph{local continuation}, it provides no explicit signal for whether a partial state is globally coherent.

\subsection{Deriving mean confidence}
\label{sec:derive_mc}

In contrast, dLLMs are trained to predict clean tokens at \emph{all} masked positions from partially filled sequences where an \emph{arbitrary} subset of the answer has been revealed, not just a prefix. 
This makes it possible to backtrack by remasking any part of a candidate's partial answer, and gives the model a more global sense of uncertainty about the current state.
This leads us to ask:
\vspace{-0.05in}
\begin{center}
    \textbf{\xblue{Does a dLLM already carry a signal of proximity to a globally coherent completion?}}
\end{center}
\vspace{-0.05in}
\textbf{Idealized Gibbs model.} To study this question more precisely, we consider an idealized Gibbs distribution, $p_{\mathrm{data}}(\rvx)\propto e^{-\beta V(\rvx)}$, where $V\colon \mathcal{V}^L \to \mathbb{R}_{\ge 0}$ is a potential function and $\beta >0 $ is an inverse-temperature parameter. 
This model provides a \emph{natural abstraction} of globally constrained data distributions: $V(\rvx)$ is defined over sequence space, and the distribution puts more probability mass into the sequence with low global potential. 
In ZebraLogic and Nurse Rostering, $V(\rvx)$ can be viewed as measuring the number of violated constraints, so the solutions correspond to low-energy states. 
In JSSP, $V(\rvx)$ encodes both feasibility and suboptimality.

Strictly speaking, the solution distribution of globally constrained problems corresponds to the zero-temperature limit $\beta\to\infty$, where the Gibbs measure is supported on satisfying or globally optimal configurations. 
We use the finite-$\beta$ Gibbs model as a smooth relaxation of this limit and as an abstraction from which to derive our notion of mean confidence.

\textbf{From the Gibbs model to mean confidence.} Assume that a masked diffusion model $f_\theta$ is trained on the idealized Gibbs distribution introduced above, i.e., $f_\theta^i(\cdot\mid \rvz)\approx p_{\mathrm{data}}(\rvx^i=\cdot\mid \rvz)$. We now ask whether these \emph{coordinate-wise posteriors} can be used as a reliability signal for the current state $\rvz$. 

To answer this, we define the \xblue{\emph{(ground-truth) mean confidence}} as follows:
\begin{equation*}
    \mathcal{C}(\rvz)\colon=\frac{1}{|M|}\sum_{i\in M} \max_v\,   p_{\mathrm{data}}(\rvx^i = v\mid \rvz),
\end{equation*}
where $M=\{i\,|\,\rvz^i=\mask\}$ is the set of masked indices. In other words, $\mathcal{C}(\rvz)$ averages how concentrated the conditional posterior is across all masked positions. This simple notion of mean confidence connects to \emph{global coherence}: write $V^\star(\rvz)=\min_{\rvx \succeq \rvz}V(\rvx)$ for the lowest potential attainable by any completion of $\rvz$. To clarify, $\rvx \succeq \rvz$ indicates sequences $\rvx$ that agree with $\rvz$ on the non-masked indices. Under suitable finite-support assumptions, 
\begin{equation*}
    \mathbb{E}_{\rvx \succeq \rvz}[V(\rvx)\,|\,\rvz] - V^\star(\rvz) \le \psi(\mathcal{C}(\rvz)) \times |M|/\beta
\end{equation*}
holds, where $\psi(c) = -c\log c -(1-c)\log(1-c)+(1-c)\log(|\mathcal{V}|-1)$ is a monotonically decreasing function over $c$.
The implication of the above inequality is that a larger $\mathcal{C}(\rvz)$ yields a stronger upper bound: the conditional expectation over all completions $\mathbb{E}_{\rvx \succeq \rvz}[V(\rvx)\,|\,\rvz]$ is closer to the best attainable potential $V^\star(\rvz)$. Since a masked diffusion model approximates the unmasking posterior, the model gives direct access to the empirical mean confidence in practice, i.e.,
\vspace{-0.05in}
\begin{equation*} \label{eq:emp-mean-conf}
    \mathcal{C}_\theta(\rvz)\colon = \frac{1}{|M|}\sum_{i\in M} \max_v\,   f_\theta^i( v\mid \rvz) \approx   \mathcal{C}(\rvz).
\end{equation*}
\vspace{-0.10in}

\textbf{Interface-level distinction from LLMs.}
The analysis above motivates mean confidence as a model-internal signal of global coherence that naturally arises from a pretrained dLLM.
% We emphasize that autoregressive models, by contrast, cannot expose an analogous signal, as they predict only the next-token distribution given a partially generated prefix $\rvx^{<i}$. 
We emphasize that autoregressive models, by contrast, do not natively expose an analogous signal, as they predict next-token distributions conditioned on partially generated prefixes $\rvx^{<i}$.
As a result, they \xblue{\textbf{cannot directly \emph{condense}}} a state-level reliability signal analogous to mean confidence from their native next-token predictions alone.

\textbf{Prior work on deploying mean confidence.}
We note that variants of mean confidence have appeared previously in the dLLM literature~\citep{lee2025lookahead,xu2025lopa}. 
However, these works use mean confidence in a distinct context, either to accelerate parallel decoding or to support particle-based decoding procedures, whereas our work leverages it for inference-time control tailored to globally structured tasks. 
We defer a detailed comparison to Section~\ref{sec:mdm_baseline}.

\subsection{Empirically verifying mean confidence}
% We now verify whether mean confidence provides a meaningful signal of global coherence in practice, particularly using our motivating examples, ZebraLogic and JSSP.
We now verify whether mean confidence provides a meaningful signal of global coherence in practice, using ZebraLogic and JSSP as representative feasibility and optimization tasks.
We first elaborate on the design of each task and then show how mean confidence appears as a signal of global coherence. 
This leads to the central algorithmic ingredients of what we call \emph{\xblue{\textbf{blackboard intelligence}}}.

\begin{figure}[h]
    \centering
    \vspace{-0.8em}
    \begin{minipage}[t]{0.45\linewidth}
        \centering
        \includegraphics[width=\linewidth]{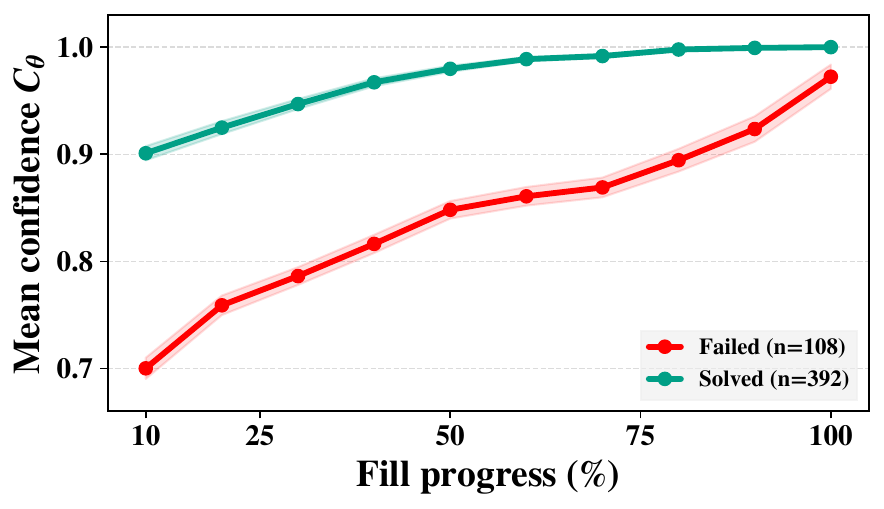}
    \end{minipage}
    \hfill
    \begin{minipage}[t]{0.53\linewidth}
        \centering
        \includegraphics[width=\linewidth]{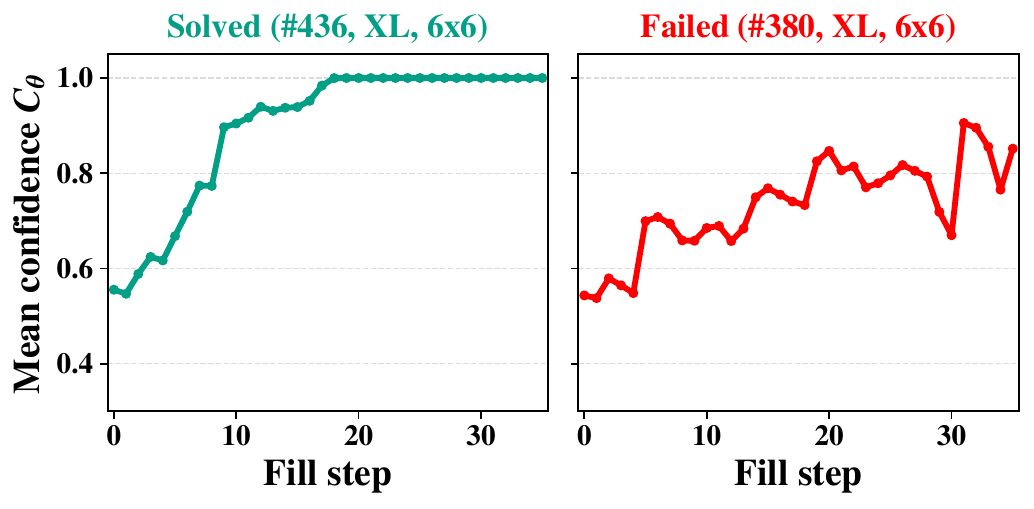}
    \end{minipage}
    \vspace{-0.12in}
    \caption{\textbf{Mean confidence trajectories on ZebraLogic.}
(\textbf{Left}) Aggregated over all evaluation instances, \coloredul{xxgreen}{solved} puzzles exhibit consistently higher mean confidence $\mathcal{C}_\theta$ than \coloredul{xred}{failed} ones.
(\textbf{Right}) Per-puzzle trajectories: mean confidence rises smoothly toward $1.0$ for correctly solved puzzles (\coloredul{xxgreen}{green}), whereas it remains unstable for incorrect ones (\coloredul{xred}{red}).}
    \label{fig:zl_mc}
    \vspace{-1.2em}
\end{figure}

\textbf{ZebraLogic.} 
Each puzzle in ZebraLogic is generated by a constraint-based generator~\citep{lin2025zebralogic} and then converted into natural-language clues. 
Following the official ZebraLogic benchmark, we group instances into
\{Small, Medium, Large, X-Large\} tiers according to log search-space size.
To further characterize constraint-solving difficulty, we encode each puzzle
with a Z3 SMT solver and record the number of conflicts encountered during
search. This conflict count provides a complementary measure of constraint
difficulty within and across the search-space tiers.
To obtain a stronger stress test, we construct ZebraLogic-Hard following the same recipe: a 500-instance held-out variant that preserves the official benchmark format while increasing constraint difficulty. ZebraLogic-Hard consequently has a broadly similar search-space scale across the four difficulty tiers, but substantially higher Z3 conflict counts. We detail the task construction in Appendix~\ref{app:zl_data}.
\footnote{For simplicity, we refer to this curated benchmark as ZebraLogic throughout the manuscript. We report additional experimental results on the original ZebraLogic task from \citet{lin2025zebralogic} in Appendix~\ref{app:zl_official}.}

For evaluation, we fine-tune LLaDA-8B-Instruct~\citep{nie2025large} on 30k training instances with LoRA~\citep{hu2021lora} and evaluate it on 500 held-out instances. We use greedy decoding as defined in Section~\ref{sec:prelim}; throughout this work, this refers to the instance that unmasks a single token at each step.

Under this evaluation setup, for each correct and incorrect puzzle, we track the mean confidence $\mathcal{C}_\theta(\rvx_{t_\cdot})$ along the intermediate inference trajectory and aggregate the trajectories across puzzles. 
As shown in Figure~\ref{fig:zl_mc} (left), we observe a clear separation trend: \xxpurple{\textbf{the mean confidence trajectory for correctly solved puzzles}} (\coloredul{xxgreen}{green line}) is \xxpurple{\textbf{\emph{consistently higher} than that for incorrectly solved puzzles}} (\coloredul{xred}{red line}).
Interestingly, we also observe that the confidence trajectory is not only higher for correctly solved puzzles but also more stable. In particular, incorrectly solved puzzles often exhibit confidence drops after wrong token fills, leading to an \xxpurple{\textbf{unstable confidence trajectory}}, whereas correctly solved puzzles maintain a \xxpurple{\textbf{more stable trajectory}} (Figure~\ref{fig:zl_mc}, right).
% Added
The same pattern holds on Nurse Rostering (Appendix~\ref{app:nr_confidence_by_difficulty}).

\begin{figure}[h]
    \centering
    \vspace{-0.5em}
    \begin{minipage}[t]{0.45\linewidth}
        \centering
        \includegraphics[width=\linewidth]{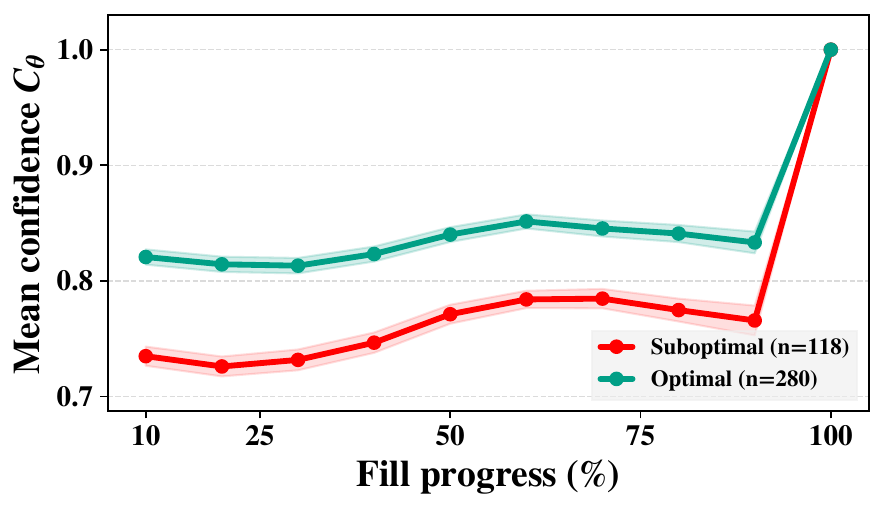}
    \end{minipage}
    \hfill
    \begin{minipage}[t]{0.523\linewidth}
        \centering
        \includegraphics[width=\linewidth]{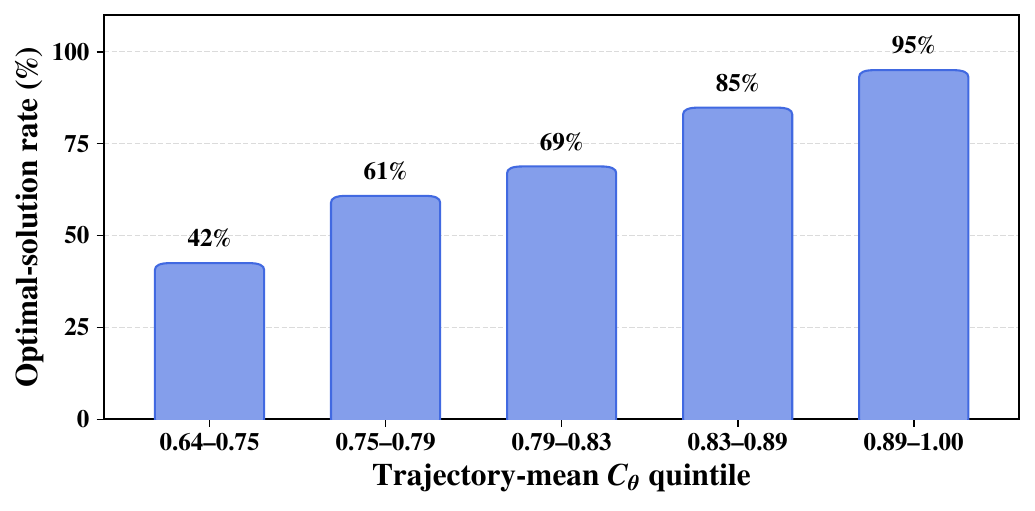}
    \end{minipage}
    \vspace{-0.12in}
    \caption{\textbf{Mean confidence trajectories on JSSP.} (\textbf{Left}) Across evaluation instances, inference trajectories that yield \coloredul{xxgreen}{optimal solutions} exhibit higher mean confidence than those that induce \coloredul{xred}{suboptimal solutions}. (\textbf{Right}) Optimal-solution rate increases monotonically across quintiles of trajectory-averaged mean confidence.}
    \label{fig:jssp_mc}
    \vspace{-0.5em}
\end{figure}

\textbf{JSSP.}
We now ask whether mean confidence remains practically useful in a more challenging domain, namely JSSP. Unlike ZebraLogic, a sequence here is evaluated by its makespan rather than simply by whether it satisfies all constraints, likely inducing a more complex posterior and confidence landscape. 
Adopting the natural-language input format of \citet{abgaryan2025starjob}, which evaluates LLMs on the JSSP, we construct 30k training examples and an evaluation set of 400 instances.
We detail the task construction in Appendix~\ref{app:jssp_data}.

We use the same greedy decoding and repeat the run-level analysis. 
In particular, we compare the aggregated mean confidence between runs that yield globally optimal makespans and runs that yield suboptimal makespans. 
As shown in Figure~\ref{fig:jssp_mc} (left), runs that produce optimal solutions maintain higher mean confidence throughout inference, with a clear separation. 
% As in ZebraLogic, we observe a connection between the stability of the mean confidence trajectory and global coherence. 
As in ZebraLogic, we observe a connection between mean confidence trajectory and global coherence.
In particular, the aggregate mean confidence along the intermediate inference trajectory is strongly associated with the optimal-solution rate, as shown in Figure~\ref{fig:jssp_mc} (right).
\vspace{-0.02in}
\section{Confidence-Driven Blackboard Inference} 
\label{sec:experiments}

In this section, we present how we intervene at inference time using the mean confidence signal and how this enables blackboard intelligence. In particular, we ask the following questions:
\vspace{-0.05in}
\begin{itemize}[leftmargin=0.8em]
    \item \xxpurple{\textbf{Section~\ref{sec:algorithm}}}: How can we utilize mean confidence to intervene at inference time?
    \item \xxpurple{\textbf{Section~\ref{sec:ar}}}: How does Blackboard inference compare with autoregressive inference?
    \item \xxpurple{\textbf{Section~\ref{sec:mdm_baseline}}}:
    How does Blackboard compare with existing dLLM methods?
\end{itemize}
\vspace{-0.06in}

We study these questions across the three domains introduced in Section~\ref{sec:challenge}; ZebraLogic, Nurse Rostering, and JSSP, with additional transfer studies in related globally constrained feasibility and optimization settings (Appendix~\ref{app:transfer}).

\vspace{-0.5em}
\subsection{Blackboard as an inference interface}
\label{sec:algorithm}

We develop inference-time algorithms around the empirical observation from Section~\ref{sec:challenge}: higher mean confidence suggests that the current trajectory is reliable, while drops in mean confidence suggest that recent fills may have moved the state away from a globally coherent completion.
Motivated by this signal, we build our inference procedures around three simple algorithmic components, adapting the corrective action to task structure only where necessary.

\vspace{-0.8em}
\begin{itemize}[leftmargin=0.7em]
\item \coloredhl{xblue}{\textbf{$\Call{Trigger}
{\{\mathcal{C}_\theta(\rvx_{t_i})\}_{i=0}^{N-1}, \rho, \tau}$}}:
Given a greedy decoding trajectory $\{\rvx_{t_i}\}_{i=0}^{N-1}$, we summarize mean confidence over a task-specific late phase to decide whether additional inference is needed.
If the resulting late-phase confidence statistic is at least $\tau$, we keep the greedy output; otherwise, we trigger additional inference.
The task-specific statistic is defined in Appendix~\ref{app:threshold}; the late-phase fraction $\rho$ and threshold $\tau$ are selected on held-out training traces and fixed before test evaluation.
This avoids spending extra computation on trajectories that appear reliable, while allocating additional inference to trajectories whose confidence becomes unreliable.
\item \coloredhl{xblue}{\textbf{$\Call{Search}{\rvz, f_\theta, d, k}$}}:
When additional inference is triggered, mean confidence can guide search over candidate continuations from the current grid $\rvz$. 
At each masked cell, we branch over the top-$k$ candidate values, simulate a depth-$d$ greedy decoding rollout per branch, and choose the branch with the highest mean confidence. 
This gives a simple way to search over the fixed canvas using the dLLM's own confidence signal.
\item \coloredhl{xblue}{\textbf{$\Call{Backtrack}{\rvz, f_\theta}$}}: We use confidence drops as a signal of coherence loss.
If a new fill decreases mean confidence, we treat it as unreliable and backtrack one step by reverting the fill.
\end{itemize}

\vspace{-0.13in}
\begin{algorithm}[H]
\small
\caption{Confidence-guided corrective inference for feasibility tasks}
\label{alg:feasibility}
\begin{algorithmic}[1]

\Require Prompt $p$, dLLM $f_\theta$, trigger parameters $(\rho,\tau)$, anchor threshold $\alpha$, search depth/width $(d,k)$

\State $(\hat{\rvz}, \{\mathcal{C}_\theta(\cdot)\}_{i=0}^{N-1})
       \gets \Call{Greedy}{p, f_\theta}$
       \Comment{\xslategray{initial greedy decoding}}

\State \textbf{if}
\coloredhl{xblue}{$\Call{Trigger}{
\{\mathcal{C}_\theta(\cdot)\}_{i=0}^{N-1},\rho,\tau}
= \mathtt{False}$}:
\textbf{return} $\hat{\rvz}$
\Comment{\xslategray{skip additional inference}}

\State $\rvz \gets \Call{Initialize}{p}$
    \Comment{\xslategray{restart from prompt}}

\While{$\rvz$ has masked cells}

    \State \textbf{if} $\mathcal{C}_\theta(\rvz) < \alpha$:
    \coloredhl{xblue}{$
    \rvz \gets \Call{Search}{\rvz,f_\theta,d,k}$}
    \Comment{\xslategray{mean-confidence-guided search}}

    \State \textbf{else}:
    \coloredhl{xblue}{$
    (\rvz_{\mathrm{next}},\mathtt{accepted})
    \gets \Call{Backtrack}{\rvz,f_\theta}$}
    \Comment{\xslategray{confidence-gated fill}}

    \State \algindent
    \textbf{if} $\mathtt{accepted}$:
    $\rvz \gets \rvz_{\mathrm{next}}$

    \State \algindent
    \textbf{else}:
    \coloredhl{xblue}{$
    \rvz \gets \Call{Search}{\rvz,f_\theta,d,k}$}
    \Comment{\xslategray{invoke search}}

\EndWhile

\State \textbf{return} $\rvz$

\end{algorithmic}
\end{algorithm}
\vspace{-0.15in}

\textbf{Constrained feasibility.}
For ZebraLogic and Nurse Rostering, the goal is to recover a globally feasible assignment satisfying all constraints, so we use mean confidence directly to guide search and revision.
We instantiate this corrective procedure in Algorithm~\ref{alg:feasibility}.
We first run greedy inference and retain its output if the confidence trigger does not fire; otherwise, we re-initialize from the prompt and alternate between $\Call{Backtrack}{\rvz, f_\theta}$ and $\Call{Search}{\rvz, f_\theta, d, k}$.
When $\mathcal{C}_\theta(\rvz) < \alpha$, we invoke search directly; otherwise, backtracking accepts only fills that do not decrease mean confidence, invoking search if no fill is accepted.
We use $\alpha=0.9$, search depth $d=3$, and width $k=5$.
For ZebraLogic, the trigger uses the late-phase minimum with $(\rho,\tau)=(0.80,1.0)$, whereas Nurse Rostering uses the late-phase mean with $(\rho,\tau)=(0.90,0.95)$.

\textbf{Global optimization.}
For JSSP, where solution quality is determined by makespan, we use mean confidence to decide when additional inference is needed and the task objective to select among completed candidates.
We first run greedy inference and apply $\Call{Trigger}{\{\mathcal{C}_\theta(\cdot)\}_{i=0}^{N-1}, \rho, \tau}$ with $(\rho,\tau)=(0.50,0.70)$.
If the trigger does not fire, we keep the greedy schedule.
Otherwise, we perform Best-of-$N$ sampling with $N=10$ and retain the valid schedule with the lowest makespan.

Across both settings, \emph{\textbf{\xblue{blackboard intelligence}}} uses state-level confidence to decide when the current canvas should be trusted or further explored.
This provides a common inference-control interface, with corrective actions that adapt to task structure: search and revision for constrained feasibility, and objective-guided candidate exploration for global optimization.

\vspace{-1.0em}
\subsection{Comparison to Autoregressive Inference}
\label{sec:ar}
\vspace{-0.05in}

In this section, we compare Blackboard with autoregressive inference on the same globally constrained problems.
We fine-tune LLaMA-3.1-8B-Instruct and LLaDA-8B-Instruct on the same $30$k task-specific examples with matched configurations, using their respective next-token and any-order prediction objectives.
Both models use the same one-cell--one-token output representation, aligning each commitment with an atomic decision variable in the underlying constrained assignment.
This setup lets us isolate the effect of Blackboard within a fixed LLaDA checkpoint while comparing against autoregressive inference after matched downstream adaptation.

\textbf{Across-domain comparison.}
Across all three domains, Blackboard consistently improves the corresponding fine-tuned LLaDA checkpoint (Figure~\ref{fig:main_result}).
On ZebraLogic, accuracy improves from $78.4\%$ to $90.4\%$; on Nurse Rostering, exact feasibility improves from $73.0\%$ to $76.4\%$; and on JSSP, optimality improves from $70.0\%$ to $80.2\%$.
These gains show that state-level confidence can support effective inference-time control across globally constrained tasks.
Detailed per-domain comparisons are provided in Appendices~\ref{app:zl_main_results}, \ref{app:nr_by_difficulty}, and \ref{app:jssp_main_results}.

Autoregressive search, however, shows a different cross-domain pattern under simple continuation search: beam search improves LLaMA on JSSP but yields only marginal gains on ZebraLogic and Nurse Rostering.
This contrast is consistent with JSSP admitting many valid or near-optimal completions, so preserving alternative trajectories can retain promising schedules, whereas ZebraLogic and Nurse Rostering require coordinated satisfaction of globally coupled constraints, leaving continuation search alone with limited opportunities to repair an inconsistent partial assignment.

\begin{wrapfigure}[18]{r}{0.47\textwidth}
    \vspace{-0.12in}
    \centering
    \includegraphics[
        width=\linewidth,
        trim={0.2cm 0.2cm 0.2cm 0.2cm},
        clip
    ]{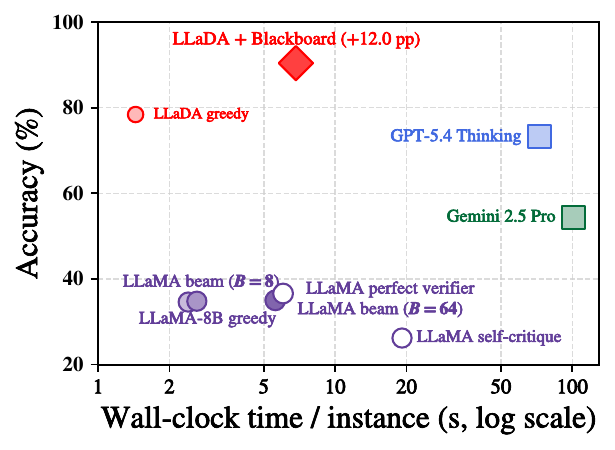}
    \caption{
    \textbf{Accuracy--latency on ZebraLogic.}
    Blackboard improves the same LLaDA-8B checkpoint, while autoregressive search and refinement show limited accuracy scaling.
    }
    \label{fig:zl_accuracy_latency}
    \vspace{-0.15in}
\end{wrapfigure}

\textbf{Scaling autoregressive inference.}
On ZebraLogic, autoregressive accuracy gains remain limited despite substantially greater inference-time computation allocated to wider search and explicit refinement, and do not match the gain achieved by Blackboard over standard inference with the same LLaDA checkpoint.
Figure~\ref{fig:zl_accuracy_latency} summarizes this scaling behavior on a common observed-latency axis.
We first scale continuation search, where widening beam search from $B=8$ to $B=64$ increases accuracy only marginally, from $34.8\%$ to $35.0\%$.
We then move beyond continuation search and give the same fine-tuned LLaMA checkpoint a persistent candidate solution with an explicit opportunity to revise earlier decisions.
In \emph{self-critique}, the model identifies clues it believes are violated and regenerates a revised solution conditioned on its own diagnosis.
In \emph{perfect-verifier refinement}, a programmatic checker instead supplies the exact violated clues before regeneration.
Self-critique reaches $26.2\%$, and even with exact error localization, perfect-verifier refinement reaches only $36.6\%$, compared with $34.6\%$ under greedy decoding.
Further details are provided in Appendix~\ref{app:zl_ar_refinement_protocol}.

The fine-tuned LLaMA and LLaDA systems are measured on the same hardware, while frontier-model latency is shown as an API-level practical reference.
Hardware-matched compute measurements are reported in Appendix~\ref{app:zl_ar_hardware_compute}.
Although autoregressive decoding is substantially more FLOP-efficient due to KV caching, the hardware-matched measurements show that this compute-efficiency advantage does not translate into stronger accuracy scaling from additional search and refinement.

\textbf{Frontier LLMs.}
We finally broaden the comparison to frontier autoregressive models that combine substantially greater model scale with strong test-time reasoning, including chain-of-thought, self-verification, and dedicated reasoning modes.
GPT-5.4 Thinking is the strongest tested configuration: Blackboard reaches $90.4\%$ versus $73.2\%$ on ZebraLogic and $80.2\%$ versus $77.0\%$ optimality on JSSP, while GPT-5.4 Thinking remains stronger on Nurse Rostering.
These comparisons place the gains from state-level inference control in the context of substantially larger models with strong general-purpose test-time reasoning.
Full prompting and latency details are provided in Appendix~\ref{app:frontier_ref}.

% Together, these comparisons suggest that Blackboard's gains arise not simply from more test-time computation, but from state-level control over a persistent, revisable candidate---an advantage that stronger autoregressive inference does not consistently reproduce.
Together, these results suggest that Blackboard provides state-level inference control that stronger autoregressive inference does not consistently recover with additional test-time computation.

\subsection{Comparison to Existing dLLM Methods and Ablations}
\label{sec:mdm_baseline}

Having contrasted Blackboard with autoregressive inference, we next situate it among existing dLLM inference-time methods and use controlled variants to examine the sources of its gains.
We consider remasking methods~\citep{wang2025remasking,kim2025fine,schiff2026learn,huang2025don}, represented by PRISM~\citep{kim2025fine}, and search-based methods~\citep{shen2026improving,xu2025lopa,fu2025bits}.
Among the latter, LoPA~\citep{xu2025lopa} is particularly relevant because it also uses mean-confidence lookahead. The key distinction lies in how the signal is used: LoPA uses it to enable parallel commitment for reducing the inference cost, whereas we use it to evaluate candidate states and determine when further inference is needed.

To isolate the role of selective triggering, we additionally evaluate an \emph{always-on} Blackboard variant that applies the same corrective inference procedure to every puzzle, without the puzzle-level confidence trigger.
Further intervention and search ablations are provided in Appendix~\ref{app:zl_ablations_baselines}.

\begin{wraptable}[9]{r}{0.45\textwidth}
\vspace{-0.14in}
\centering
\caption{\textbf{dLLM inference on ZebraLogic.}}
\label{tab:dllm_inference}

\fontsize{8.5pt}{9.5pt}\selectfont
\renewcommand{\arraystretch}{0.92}

\begin{tabularx}{\linewidth}{Xrr}
\toprule
\textbf{Inference} & \textbf{NFE} & \textbf{Acc. (\%)} \\
\midrule
Greedy                 & 35   & 78.4 \\
PRISM                  & 54   & 79.2 \\
LoPA                   & 6.74 & 73.0 \\
Blackboard (always-on) & 280  & 86.8 \\
\rowcolor{xxpurple!20}
\textbf{Blackboard}    & \textbf{129} & \textbf{90.4} \\
\bottomrule
\end{tabularx}

\vspace{-0.10in}
\end{wraptable}

Table~\ref{tab:dllm_inference} reveals distinct uses of confidence during dLLM inference on Zebralogic.
PRISM modestly improves over greedy decoding ($79.2\%$ versus $78.4\%$), suggesting that token-level reconsideration alone provides limited gains on ZebraLogic.
In contrast, LoPA substantially reduces inference effort, from $35$ NFE under greedy decoding to $6.74$ NFE, illustrating how mean-confidence lookahead can support efficient parallel commitment.
Applying Blackboard correction to every puzzle already improves accuracy to $86.8\%$.
Selective confidence triggering further raises accuracy to $90.4\%$ while reducing average inference effort from $280$ to $129$ NFE, showing that state-level confidence is useful not only for guiding corrective inference but also for deciding when such intervention should be invoked.

This pattern extends across domains, with the role of confidence adapting to task structure.
On Nurse Rostering, selective triggering improves over both greedy inference and always-on correction ($76.4\%$ versus $73.0\%$ and $57.4\%$; Appendix~\ref{app:nr_trigger_compute}).
On JSSP, confidence instead allocates objective-based candidate exploration, reducing average NFE from $217$ to $149$ relative to untriggered BoN while retaining comparable solution quality (Appendix~\ref{app:jssp_trigger_compute}).
Together, these results support state-level confidence as a reusable inference-control signal across globally constrained problems.
\section{Conclusion}
\label{sec:conclusion}

Recent rapid progress in generative modeling, especially in LLMs, has largely centered on scaling next-token prediction and improving post-training.
Our results highlight a complementary axis of progress: the \emph{inference interface} itself can shape how models solve globally constrained problems.
Diffusion language models, equipped with \emph{blackboard intelligence}, expose partially specified solution states that can be evaluated, searched, and revised throughout inference, with state-level confidence providing a native signal for controlling this process.

Across three representative globally constrained settings---ZebraLogic, Nurse Rostering, and JSSP---our analysis shows that mean confidence tracks the global coherence of partially specified solutions.
Building on this signal, Blackboard uses state-level confidence to control when and how inference intervenes, consistently improving inference while holding the fine-tuned LLaDA-8B checkpoint fixed and outperforming same-scale autoregressive baselines.
Moreover, stronger autoregressive inference does not consistently recover these gains with additional test-time computation.

Together, our results suggest that progress on globally constrained problem solving may depend not only on larger models or stronger deliberation, but also on \emph{what intermediate state a model exposes to inference} and how that state can be evaluated and revised.
Our study focuses on globally constrained problems, where success depends on maintaining coherence across interacting decisions.
An important next question is how far \emph{blackboard intelligence} extends beyond this regime and which problems benefit most from revisable inference over partial solution states.

\newpage
\bibliographystyle{plainnat}
\bibliography{main}
\appendix
\newpage
\section{Related Works}
\label{sec:related}
\paragraph{Globally constrained reasoning.}
Recent work has increasingly evaluated language models on tasks whose correctness depends on global feasibility, consistency, or optimality rather than local textual plausibility. \emph{ZebraLogic} identifies a ``curse of complexity'' in logic-grid reasoning, where accuracy degrades sharply as the underlying constraint structure becomes harder~\citep{lin2025zebralogic}. \emph{Enigmata} broadens this direction to synthetic verifiable puzzles with controllable difficulty and rule-based evaluation~\citep{chen2026enigmata}, while \emph{SATBench} and \emph{REL} study assignment search and relational reasoning under many interacting constraints~\citep{wei2025satbench,fesser2026evaluating}. For optimization, \emph{StarJob}, \emph{NLCO}, and \emph{ConstraintBench} evaluate whether LLMs can directly produce feasible or near-optimal solutions to scheduling and combinatorial optimization problems~\citep{abgaryan2025starjob,jiang2026reasoning,tso2026constraintbench}.
Our experiments focus on three representative settings: ZebraLogic for unique constrained feasibility, Nurse Rostering for operational constrained feasibility, and JSSP for global optimization.

\paragraph{Diffusion language models and comparison to autoregressive models.}
Following the advent of masked diffusion models, they have been scaled up~\citep{nie2025large,song2025seed,gong2025diffucoder}, showing benchmark performance comparable to their LLM counterparts. These models commonly employ \emph{confidence-based decoding} as the default inference-time algorithm. Several preliminary lines of work show that inference-time flexibility, instantiated by any-order generation, can induce behavior fundamentally different from next-token prediction~\citep{nagarajan2025roll,kim2025train,ye2024beyond,anil2025interleaved}. This has also motivated a line of research on inference-time control.

\paragraph{Inference-time control for diffusion language models.} A line of remasking work begins with \citet{wang2025remasking}, where positions are heuristically chosen for remasking to revise some early decisions. Later work improves this by learning where to remask~\citep{kim2025fine,huang2025don,schiff2026learn}. Although these methods differ slightly in formulation and empirical consequences, they are all based on per-position likelihood, which may lack a global coherence signal.

Another growing line of work studies inference-time search using additional signals. \citet{lee2025lookahead} use mean confidence as a quantity for simulating sequential Monte Carlo. \citet{xu2025lopa} employ mean confidence as the main search quantity in a branching-based search, although they report gains mainly in inference-time speed-up by enhancing parallel decoding ability. \citet{fu2025bits,shen2026improving} also employ search algorithms, but differ in both motivation and the quantities used.

In summary, prior work points to the limitations of LLMs on globally structured problems, as well as the use of mean confidence to drive inference-time algorithms for dLLMs. In contrast, our work takes a distinct perspective: we use mean confidence as a proxy for global coherence, deciding when to trust a trajectory, when to search, and when to backtrack, without changing the architecture or training objective.

% jay: we should be more careful in our arxiv version.
% ============================================================
% GIBBS DERIVATION
% ============================================================
\section{Mean Confidence under the Idealized Gibbs Model}
\label{app:gibbs}

We provide the derivation underlying the bound stated in
Section~\ref{sec:derive_mc}.
Consider a Gibbs distribution
\[
p_{\mathrm{data}}(\rvx) = \frac{1}{Z}\, e^{-\beta V(\rvx)},
\qquad
\rvx \in \mathcal{V}^L,
\]
where $V : \mathcal{V}^L \to \mathbb{R}_{\geq 0}$ is a non-negative potential and $\beta > 0$ is an inverse-temperature parameter. For a partial state $\rvz$ with masked positions $M = \{i : \rvz^i = \mask\}$, write
\[
V^\star(\rvz) \;=\; \min_{\rvx \succeq \rvz} V(\rvx),
\]
where the minimum is over all completions of $\rvz$.

We bound the conditional excess potential $\mathbb{E}[V(\rvx) \mid \rvz] - V^\star(\rvz)$ in terms of the (true) mean confidence
\[
\mathcal{C}(\rvz)
=
\frac{1}{|M|}
\sum_{i \in M}
\max_v p_{\mathrm{data}}(\rvx^i = v \mid \rvz).
\]
The argument proceeds in two steps.

\paragraph{Step 1: entropy upper-bounds excess potential.}
Let $q(\rvx)=p_{\mathrm{data}}(\rvx\mid \rvz)$ be the conditional distribution over completions of $\rvz$. Since $V(\rvx)\geq V^\star(\rvz)$ for all completions and $q(\rvx) \propto e^{-\beta V(\rvx)}$, standard Gibbs variational manipulation gives
\[
\mathbb{E}_{q}\!\left[V(\rvx)\right] - V^\star(\rvz)
\;\leq\;
\frac{1}{\beta}\, H(q),
\]
where $H(q)$ is the entropy of the conditional distribution over completions.

\paragraph{Step 2: Fano-type coordinate bound.}
For each masked coordinate $i\in M$, define
\[
c_i
=
\max_v p_{\mathrm{data}}(\rvx^i = v \mid \rvz).
\]
A Fano-type bound gives
\[
H(\rvx^i\mid \rvz)
\;\leq\;
\psi(c_i),
\qquad
\psi(c)
=
-c\log c -(1-c)\log(1-c) + (1-c)\log(|\mathcal{V}|-1).
\]
By subadditivity of entropy,
\[
H(q)
=
H(\rvx^M \mid \rvz)
\;\leq\;
\sum_{i\in M} H(\rvx^i\mid \rvz)
\;\leq\;
\sum_{i\in M} \psi(c_i).
\]
Since $\psi$ is concave on $[1/|\mathcal{V}|,1]$, Jensen's inequality yields
\[
\sum_{i \in M} \psi(c_i)
\;\leq\;
|M|\,\psi\!\left(\mathcal{C}(\rvz)\right).
\]
Combining the two steps,
\[
\mathbb{E}\!\left[V(\rvx) \mid \rvz\right] - V^\star(\rvz)
\;\leq\;
\frac{|M|}{\beta}\,
\psi\!\left(\mathcal{C}(\rvz)\right),
\]
which recovers the bound in Section~\ref{sec:derive_mc}. Because $\psi$ is monotone decreasing on $[1/|\mathcal{V}|,1]$, high mean confidence forces this upper bound on conditional excess potential to be small.

\paragraph{Caveat.}
The bound is stated for $\mathcal{C}$ computed under the true conditional $p_{\mathrm{data}}(\,\cdot \mid \rvz)$, which is not available in practice. Our experiments compute the model-derived $\mathcal{C}_\theta(\rvz)$ from an SFT-trained dLLM, as defined in Section~\ref{sec:derive_mc}; the derivation motivates this choice but does not directly bound the model-derived quantity.

% ============================================================
% EXPERIMENTAL SETUP
% ============================================================
\section{Experimental Setup}
\label{app:setup}

\subsection{Tasks and Data}
\label{app:data}

We follow the data formulation in Section~\ref{sec:challenge}. 
This appendix details construction choices for our setup.

\subsubsection{ZebraLogic}
\label{app:zl_data}

\paragraph{Output representation.}
We restrict the attribute-value vocabulary to entries that tokenize as a single token under the LLaMA and LLaDA tokenizers, fixing the unit of commitment at one cell = one token. 
This isolates decoding behavior from multi-token surface-form effects (sub-token boundaries, tokenizer-specific splits, copy-from-prompt heuristics) so that the main source of difficulty is the constraint structure rather than surface-form tokenization.

\paragraph{Generation and verification.}
We adapt the Enigmata puzzle generator~\citep{chen2026enigmata} with three modifications: single-token vocabulary, higher Z3 conflict targets, and ZebraLogic-style natural-language clue templates.
Each instance is produced by a constraint-based generator that first samples a target grid of size $N \times M$ with single-token attribute values, then emits clues from a fixed predicate vocabulary: \texttt{eq}, \texttt{ne}, \texttt{directly\_left/right}, \texttt{somewhere\_left/right}, \texttt{next\_to}, \texttt{far\_left/right}, \texttt{between}, \texttt{n\_houses\_between}, \texttt{at\_house\_N}, and \texttt{not\_at\_house\_N}. 
We verify uniqueness with the Z3 SMT solver~\citep{de2008z3}; instances admitting multiple solutions or failing within the solver budget are rejected.

\paragraph{Difficulty tiers.}
Tier assignment follows the official ZebraLogic binning by log search-space size~\citep{lin2025zebralogic}. 
Relative to the official benchmark, ZebraLogic-Hard has a broadly similar search-space scale but substantially higher Z3 conflict counts across difficulty tiers (Table~\ref{tab:zl_dataset}), making it a stronger stress test of partial-assignment feasibility.

\begin{table}[h]
\centering
\caption{ZebraLogic-Hard vs.\ the official ZebraLogic benchmark.
Search-space sizes are comparable across tiers, while Z3 conflict counts are higher in our variant. 
Z3 conflicts are averaged over $32$ runs with different random seeds.}
\label{tab:zl_dataset}
\footnotesize
\setlength{\tabcolsep}{6pt}
\begin{tabular}{@{}l cc cc@{}}
\toprule
 & \multicolumn{2}{c}{ZL official} & \multicolumn{2}{c}{ZL-Hard} \\
\cmidrule(lr){2-3} \cmidrule(lr){4-5}
Category & log\_ss & Z3 & log\_ss & Z3 \\
\midrule
Small   & 1.26  & 0.3  & 2.37  & 6.0   \\
Medium  & 3.58  & 8.0  & 4.35  & 19.8  \\
Large   & 6.83  & 27.1 & 7.45  & 50.2  \\
X-Large & 12.38 & 83.6 & 12.60 & 161.4 \\
\midrule
Overall & 5.92  & 28.9 & 6.69  & 59.4  \\
\bottomrule
\end{tabular}
\end{table}

\paragraph{Splits and deduplication.}
The training set contains $30{,}000$ puzzles, stratified across the four difficulty tiers. The held-out evaluation set contains $500$ puzzles ($125$ per tier). 
Train and evaluation sets are generated with different random seeds and checked for exact duplicates using a fingerprint over the clue set and target grid; no duplicates were 
found.

\paragraph{Evaluation.}
A puzzle is counted as solved iff every cell in the predicted grid matches the unique satisfying assignment. There is no partial credit.

\paragraph{Official benchmark conversion.}
For single-token grid filling, category and value names from the official benchmark are mapped to single-token aliases. 
The relational predicates, constraint graph, and unique satisfying assignment are preserved. 
The adapted official ZebraLogic benchmark used in Appendix~\ref{app:zl_official} is produced by the same conversion procedure.

\subsubsection{Nurse Rostering}
\label{app:nr_data}

\paragraph{Task and output representation.}
Each instance specifies a staff-by-day roster in which every cell receives one
shift from the single-token vocabulary
$\{\texttt{day}, \texttt{late}, \texttt{night}, \texttt{off}\}$.
The prompt provides natural-language rules over staff--day assignments, and
the model outputs a Markdown table with one row per staff member and one
column per day.  As in ZebraLogic, one grid cell corresponds to one token
under both the LLaMA and LLaDA tokenizers.

\paragraph{Generation and verification.}
We generate instances by sampling a target roster and a set of relational,
temporal, and counting constraints.  Relational constraints include fixed,
equal, and unequal assignments; temporal constraints specify allowed or
forbidden consecutive shifts; and counting constraints specify row-level
shift counts, day-level coverage, or maximum consecutive work periods.
We use Z3 to verify that every retained instance has a unique satisfying
roster.

\paragraph{Difficulty bands.}
We partition the $500$ held-out instances into four equal-size difficulty
bands using the number of conflicts encountered by Z3 during constraint
solving: d0 contains $0$--$1$ conflicts, d1 contains $2$--$4$, d2 contains
$5$--$9$, and d3 contains $10$--$31$ conflicts.  This difficulty measure is
used only for reporting and is not provided to the model.

\paragraph{Splits and evaluation.}
The training set contains $30{,}000$ generated rosters.  The held-out
evaluation set contains $500$ instances, with $125$ instances in each
difficulty band.  Training and evaluation instances are generated with
different random seeds and checked for duplicate constraint sets and target
rosters.  We report exact feasibility: an output is correct only if every
predicted staff--day assignment matches the unique satisfying roster.

\subsubsection{JSSP}
\label{app:jssp_data}

\paragraph{Output representation.}
Inputs follow the natural-language Starjob description~\citep{abgaryan2025starjob}. 
For the output, instead of predicting per-operation start times, we use a $K \times J$ grid whose row $m$ lists the $J$ jobs that pass through machine $m$, in execution order. 
A grid is row-valid iff each row is a permutation of $\{0,\ldots,J-1\}$. 
For row-valid grids, we recover the induced schedule by earliest-start simulation under the specified job precedence constraints and machine queue orders, then compute its makespan. 
This compact form keeps the output length and per-cell single-token structure aligned with the ZebraLogic setup, while validity and makespan remain exactly computable from the grid.
 
\paragraph{Generation and verification.}
Each instance is generated by sampling job processing times and machine permutations, then solving the resulting job-shop instance with OR-Tools CP-SAT~\citep{ortools}. 
Instances without an optimality certificate are rejected and resampled. 
The solved schedule is converted to the compact machine-by-slot grid by sorting operations on each machine by start time; each cell stores a single-token job identifier encoded as a digit string.

\paragraph{Multiple optimal solutions.}
JSSP instances often admit multiple schedules with the same optimal makespan. 
During training-set generation we collect additional schedules that match the CP-SAT-certified optimum when available. 
During SFT, when alternative optimal schedules exist, we use them as targets with probability $0.5$ by sampling uniformly from the stored alternative set. 
At evaluation time, outputs are scored by validity and makespan rather than by cellwise agreement with a particular target grid.

\paragraph{Splits and deduplication.}
The training set contains $30{,}000$ instances. The held-out evaluation set contains $400$ instances spanning grid sizes from $3{\times}3$ through $8{\times}8$, all with CP-SAT-certified optimal makespans. 
Train and evaluation sets are generated with different random seeds and checked for exact overlap by hashing job processing times and machine permutations; no overlap was found.

\paragraph{Evaluation.}
The main text reports \% optimal and \% within $5\%$. 
Concretely, for row-valid grids we simulate the induced schedule and compute
\[
\mathrm{ms\_ratio} = 
\frac{\mathrm{actual\_makespan}}{\mathrm{optimal\_makespan}},
\]
counting a puzzle as optimal when $\mathrm{ms\_ratio} = 1$ and within $5\%$ when $\mathrm{ms\_ratio} \leq 1.05$. 
For finer-grained analysis, this appendix additionally reports mean $\mathrm{ms\_ratio}$ alongside the two main-text metrics.
Invalid outputs include row-permutation violations and row-valid grids whose induced precedence-plus-queue dependency graph contains a cycle (no feasible schedule exists). 
Invalid outputs are not counted toward the optimality or within-$5\%$ rates: \% optimal and \% within $5\%$ use the full $n{=}400$ denominator with invalid outputs counted as fail; mean $\mathrm{ms\_ratio}$ is computed over valid entries only, avoiding an arbitrary penalty value.

\subsection{SFT Recipe, Hyperparameters, and Compute}
\label{app:sft_recipe}

We fine-tune LLaMA-3.1-8B-Instruct (LLM) and LLaDA-8B-Instruct
(dLLM) on each task using matched LoRA configurations, optimizer settings,
and learning-rate schedules within a task.  Sequence length, batch
configuration, and training duration are adjusted to accommodate the
task-specific input-output format.

\begin{table}[h]
\centering
\caption{SFT hyperparameters.  LLM and dLLM configurations are matched
within each task except for architecture-required masking and decoding
mechanics.}
\label{tab:sft_configs}
\small
\setlength{\tabcolsep}{4.5pt}
\begin{tabular}{@{}lccc@{}}
\toprule
& \textbf{ZebraLogic} & \textbf{Nurse Rostering} & \textbf{JSSP} \\
\midrule
\textbf{Base models (LLM / dLLM)}
& \multicolumn{3}{l}{LLaMA-3.1-8B-Instruct / LLaDA-8B-Instruct} \\
\midrule
\textbf{LoRA} & & & \\
\quad Rank $r$ / $\alpha$ / dropout
& \multicolumn{3}{l}{$256$ / $256$ / $0.05$} \\
\quad Targets
& \multicolumn{3}{l}{q/k/v/o\_proj} \\
\midrule
\textbf{Optimization} & & & \\
\quad Optimizer (weight decay $0.01$)
& \multicolumn{3}{l}{AdamW} \\
\quad Learning rate (cosine, warmup $100$)
& \multicolumn{3}{l}{$1{\times}10^{-4}$, minimum ratio $0.1$} \\
\quad Maximum gradient norm
& \multicolumn{3}{l}{$1.0$} \\
\quad Seed
& \multicolumn{3}{l}{$2026$} \\
\midrule
\textbf{Task-specific} & & & \\
\quad Maximum sequence length
& $512$ & $1152$ & $768$ \\
\quad Micro-batch / gradient accumulation
& $6 \times 4$ & $1 \times 16$ & $3 \times 8$ \\
\quad Training schedule
& $20$ epochs & $12$ epochs & $50$ epochs \\
\quad Reported checkpoint
& final epoch & epoch $8$ (validation-best) & final epoch \\
\bottomrule
\end{tabular}
\end{table}

For Nurse Rostering, both reported LLaMA and LLaDA systems use the
epoch-$8$ checkpoint; the LLaDA checkpoint is selected using a held-out
validation split before test evaluation.

\paragraph{Compute.}
Fine-tuning was conducted on NVIDIA RTX PRO 6000 Blackwell GPUs with
96\,GB memory per GPU using PyTorch DDP.  Since sequence length and training
duration differ substantially across tasks, we do not interpret training
wall-clock time as a cross-task efficiency comparison.  Test-time NFE is
reported separately in the main results tables.

\subsection{Pre-Fine-Tuning Task Adaptation}
\label{app:raw_base}

We evaluate the instruction-tuned models before task-specific fine-tuning, using the same task inputs and task-specific output interfaces as in the matched-model comparison.
Exact task success is near zero across all three domains, showing that task-specific SFT is needed to adapt both model families to these specialized constrained output formats before their inference interfaces are compared.

\begin{table}[h]
\centering
\caption{Performance before task-specific fine-tuning. 
ZebraLogic-Hard and Nurse Rostering report exact feasibility; JSSP reports
exact optimality.}
\label{tab:raw_base}
\small
\setlength{\tabcolsep}{7pt}
\begin{tabular}{@{}l c cc@{}}
\toprule
Task & Metric & LLaDA-8B-Instruct & LLaMA-3.1-8B-Instruct \\
\midrule
ZebraLogic-Hard ($n=500$)
& exact feasibility & $0.8$ ($4/500$) & $0.4$ ($2/500$) \\
Nurse Rostering ($n=500$)
& exact feasibility & $0.0$ ($0/500$) & $0.0$ ($0/500$) \\
JSSP ($n=400$)
& exact optimality & $0.0$ ($0/400$) & $0.0$ ($0/400$) \\
\bottomrule
\end{tabular}
\end{table}

For JSSP, LLaMA produces a valid schedule on $55/400$ instances ($13.8\%$) before task-specific fine-tuning, but none is exactly optimal; LLaDA produces no valid schedules.

\subsection{Prompts}
\label{app:prompts}

All methods use the same task-specific Markdown-table output format.
Standard, thinking, and SFT rows use the base prompt. CoT rows prepend a single worked example and append an explicit reasoning and verification instruction.

\subsubsection{Base prompt}
\label{app:base_prompt}

\begin{promptbox}
### SYSTEM:
You are a precision logic solver engine.
Output ONLY a valid Markdown table as the solution.

### PUZZLE CONTEXT:
<puzzle text>

### FINAL SOLUTION:
\end{promptbox}

ZebraLogic prompts list categories, value pools, clues, and an empty Markdown
table headed by \verb!| House | <attr> | ... |!.  Nurse Rostering prompts
specify the staff members, days, shift vocabulary
(\texttt{day}, \texttt{late}, \texttt{night}, \texttt{off}), and natural-language
roster rules, followed by an empty table headed by
\verb!| Staff | D1 | ... |!.  JSSP prompts follow the Starjob-style
description~\citep{abgaryan2025starjob} and end with an empty machine-by-slot
Markdown table headed by \verb!| Machine | Slot_1 | ... |!.

\subsubsection{ZebraLogic CoT prompt}
\label{app:zl_cot_prompt}

The ZebraLogic CoT prompt prepends a worked example and appends an explicit instruction. The final answer is required to use the same Markdown-table format as the base prompt.

\paragraph{Prepended example.}

\begin{promptbox}
# Example Problem

There are 3 houses (numbered 1, 2, 3). Each house has people with two attribute categories:
  - Color: red, green, blue
  - Drink: tea, coffee, milk
Each value is unique per category and per house.

Clues:
1. The red house is at position 1.
2. The person who drinks tea is in the green house.
3. The person who drinks coffee is directly right of the milk drinker.

# Reasoning (Example)

Step 1. From clue 1: house 1 is red.
Step 2. Remaining colors {green, blue} go to houses 2, 3.
Step 3. From clue 3: coffee is directly right of milk, so milk-coffee sits at adjacent positions (1,2) or (2,3).
Step 4. From clue 2: tea is in the green house. Test case A: green at house 2. Then house 3 is blue. Drinks: tea is at house 2 (green), so milk and coffee go to houses 1 and 3. From clue 3 they must be adjacent, contradiction.
Step 5. Therefore green is at house 3. House 2 is blue. Tea is at house 3 (green). Milk-coffee adjacent: (1,2). House 1: red, milk. House 2: blue, coffee. House 3: green, tea.

# Self-check (Example)

Verify each clue against the candidate:
- Clue 1: red house at position 1 v
- Clue 2: tea drinker is in green house (house 3) v
- Clue 3: coffee (house 2) directly right of milk (house 1) v
All clues satisfied.

# Final Solution (Example)

| House | Color | Drink |
|---|---|---|
| 1 | red   | milk   |
| 2 | blue  | coffee |
| 3 | green | tea    |
\end{promptbox}

\paragraph{Appended instruction.}

\begin{promptbox}
Solve this puzzle in three sections:

1. Reasoning. Walk through the deductions step by step, considering constraints and eliminating possibilities. Use the same style as the example.

2. Self-check. Before finalizing, verify your candidate solution against EACH clue one by one. List each clue and confirm it is satisfied. If any clue is violated, revise your reasoning and produce a corrected candidate, then re-check.

3. Final Solution. Output the final answer as a Markdown table only, in the same format as the example. Do not add any commentary after the table.
\end{promptbox}

\subsubsection{Nurse Rostering CoT prompt}
\label{app:nr_cot_prompt}

The Nurse Rostering CoT prompt uses the same reasoning--self-check--final-table
structure as ZebraLogic, with a roster-specific worked example.

\paragraph{Prepended example.}

\begin{promptbox}
# Example Problem

There are 2 staff members (S1 to S2) and 3 days (D1 to D3).
Shifts: day, late, night, off.

Rules:
1. S1 works day on D1.
2. On D1, S2's shift is different from S1's.
3. For S1, a day shift is never immediately followed by a day shift.
4. S2 works off on exactly 1 day.

# Reasoning (Example)

Step 1. Rule 1 -> S1,D1 = day.
Step 2. Rule 2 -> S2,D1 != day. Rule 3 -> S1,D2 != day.
Step 3. Assign remaining cells consistently and check the counting rule
(Rule 4: S2 off exactly once).
A consistent completion is S1 = [day, night, day], S2 = [night, off, late].

# Self-check (Example)

- Rule 1: S1,D1 = day v
- Rule 2: S2,D1 (night) != S1,D1 (day) v
- Rule 3: S1,D1=day, S1,D2=night (not day) v
- Rule 4: S2 off on D2 only = 1 day v

# Final Solution (Example)

| Staff | D1 | D2 | D3 |
| --- | --- | --- | --- |
| S1 | day | night | day |
| S2 | night | off | late |
\end{promptbox}

\paragraph{Appended instruction.}

\begin{promptbox}
Solve this puzzle in three sections:

1. Reasoning. Work through the deductions step by step.

2. Self-check. Verify your candidate against EACH rule one by one. If any
rule is violated, revise and re-check.

3. Final Solution. Output ONLY the Markdown table, with one row per staff
member and one column per day. Do not add commentary after the table.
\end{promptbox}

\subsubsection{JSSP CoT prompt}
\label{app:jssp_cot_prompt}

The JSSP CoT prompt prepends a worked example and appends the instruction below. We omit clue-by-clue self-checking because JSSP has no direct analogue of natural-language clue verification.

\paragraph{Prepended example.}

\begin{promptbox}
# Example Problem

You have 3 jobs to schedule on 2 machines (Machine 0, Machine 1). Each job must visit every machine exactly once, in a fixed order. No two jobs can use the same machine at the same time. Goal: minimize the total completion time (makespan).

Job 0 requires: Machine 0 for 3 time units, Machine 1 for 2 time units.
Job 1 requires: Machine 1 for 2 time units, Machine 0 for 1 time unit.
Job 2 requires: Machine 0 for 2 time units, Machine 1 for 3 time units.
Fill in the processing order for each machine.
Each row lists the jobs processed on that machine, in order.

## Solution

| Machine | Slot_1 | Slot_2 | Slot_3 |
|---|---|---|---|
| M0 | 2 | 0 | 1 |
| M1 | 1 | 2 | 0 |
\end{promptbox}

\paragraph{Appended instruction.}

\begin{promptbox}
Think step by step about the scheduling constraints and job precedence, showing your work. Then output the final solution table in Markdown format at the end.
\end{promptbox}

% ============================================================
% ZEBRALOGIC ADDITIONAL RESULTS
% ============================================================
\section{ZebraLogic Additional Results}
\label{app:zl_additional}

\subsection{Main Evaluation Results}
\label{app:zl_main_results}

Table~\ref{tab:zl_main} reports the full comparison on ZebraLogic-Hard, including frontier LLMs, same-scale autoregressive baselines, standard LLaDA inference, and existing dLLM inference-time methods.
The main text focuses on the corresponding interface-level comparisons and inference-time diagnostics.

\begin{table}[h]
\centering
\caption{Evaluation results on ZebraLogic-Hard.}
\label{tab:zl_main}

\fontsize{9.0pt}{10.0pt}\selectfont
\renewcommand{\arraystretch}{0.87}
\setlength{\heavyrulewidth}{0.45pt}
\setlength{\lightrulewidth}{0.30pt}
\setlength{\cmidrulewidth}{0.30pt}
\setlength{\tabcolsep}{7.5pt}

\begin{tabular}{@{}lllrccccc@{}}
\toprule
\textbf{Class} & \textbf{Model} & \textbf{Inference} & \textbf{NFE}
& \textbf{S} & \textbf{M} & \textbf{L} & \textbf{XL} & \textbf{All} \\
\midrule

\multirow{4}{*}{
\coloredhl{xxpurple}{\textbf{Frontier LLMs}}
}
& \multicolumn{2}{l}{GPT-5.4}
& --- & 60.8 & 27.2 & 9.6 & 0.0 & 24.4 \\
& \multicolumn{2}{l}{GPT-5.4 (CoT prompting)}
& --- & 66.4 & 50.4 & 21.6 & 3.2 & 35.4 \\
& \multicolumn{2}{l}{GPT-5.4 Thinking}
& --- & 76.8 & 84.8 & 72.8 & 58.4 & 73.2 \\
& \multicolumn{2}{l}{Gemini 2.5 Pro}
& --- & 88.8 & 74.4 & 46.4 & 8.0 & 54.4 \\

\midrule

\multirow{4}{*}{
\coloredhl{xxpurple}{\textbf{8B-LLM}}
}
& LLaMA-3.1-8B & Greedy
& 80 & 80.8 & 48.0 & 8.8 & 0.8 & 34.6 \\
& LLaMA-3.1-8B & Beam search
& 643 & 81.6 & 48.0 & 8.8 & 0.8 & 34.8 \\
& LLaMA-3.1-8B & Self-critique
& 583 & 68.0 & 33.6 & 3.2 & 0.0 & 26.2 \\
& LLaMA-3.1-8B & Perfect verifier
& 200 & 82.4 & 51.2 & 11.2 & 1.6 & 36.6 \\

\midrule

\multirow{5}{*}{
\coloredhl{xxpurple}{\textbf{8B-dLLM}}
}
& LLaDA-8B & Greedy
& 35 & 100.0 & 94.4 & 68.0 & 51.2 & 78.4 \\

& LLaDA-8B & PRISM
& 54 & 99.2 & 94.4 & 76.0 & 47.2 & 79.2 \\

& LLaDA-8B & LoPA
& 6.74 & 100.0 & 93.6 & 61.6 & 36.8 & 73.0 \\

& LLaDA-8B & Blackboard Search
& 273 & 100.0 & 99.2 & 87.2 & 74.4 & 90.2 \\

\rowcolor{xxpurple!20}
\cellcolor{white}
& \textbf{LLaDA-8B} & \textbf{Blackboard}
& \textbf{129}
& \textbf{100.0}
& \textbf{99.2}
& \textbf{87.2}
& \textbf{75.2}
& \textbf{90.4} \\

\bottomrule
\end{tabular}
\end{table}

\subsection{Separation and Grid-Size Controls}
\label{app:zl_separation}

We report two controls for the ZebraLogic confidence-separation analysis in Section~\ref{sec:challenge}: separation across $10\%$ fill intervals and separation within fixed grid sizes. 
Solved runs maintain higher $\mathcal{C}_\theta$ throughout most of decoding (Table~\ref{tab:separation_full}), and the gap remains positive within every grid size (Table~\ref{tab:gridsize_full}).

\begin{table}[h]
\centering
\caption{Progressive separation of $\mathcal{C}_\theta$ trajectories at $10\%$-fill intervals on ZebraLogic-Hard ($n_{\text{solved}}{=}392$, $n_{\text{failed}}{=}108$).}
\label{tab:separation_full}
\small
\begin{tabular}{@{}lcccc@{}}
\toprule
Fill \% & Solved $\mathcal{C}_\theta$ & Failed $\mathcal{C}_\theta$ & $\Delta$ & Cohen's $d$ \\
\midrule
10\%  & $0.893 \pm 0.136$ & $0.682 \pm 0.093$ & $+0.211$ & 1.81 \\
20\%  & $0.913 \pm 0.122$ & $0.734 \pm 0.100$ & $+0.180$ & 1.61 \\
30\%  & $0.937 \pm 0.103$ & $0.773 \pm 0.096$ & $+0.164$ & 1.64 \\
40\%  & $0.956 \pm 0.086$ & $0.804 \pm 0.089$ & $+0.152$ & 1.73 \\
50\%  & $0.975 \pm 0.063$ & $0.835 \pm 0.087$ & $+0.140$ & 1.84 \\
60\%  & $0.984 \pm 0.048$ & $0.851 \pm 0.091$ & $+0.133$ & 1.84 \\
70\%  & $0.990 \pm 0.039$ & $0.874 \pm 0.088$ & $+0.116$ & 1.71 \\
80\%  & $0.995 \pm 0.032$ & $0.899 \pm 0.094$ & $+0.097$ & 1.38 \\
90\%  & $0.998 \pm 0.018$ & $0.916 \pm 0.108$ & $+0.082$ & 1.06 \\
100\% & $1.000 \pm 0.000$ & $0.972 \pm 0.118$ & $+0.028$ & 0.33 \\
\bottomrule
\end{tabular}
\end{table}

\begin{table}[h]
\centering
\caption{Grid-size-controlled separation of mean trajectory $\mathcal{C}_\theta$ on ZebraLogic-Hard. Separation is significant at every grid size.}
\label{tab:gridsize_full}
\small
\begin{tabular}{@{}lcccc@{}}
\toprule
Grid cells & Solved $\mathcal{C}_\theta$ & Failed $\mathcal{C}_\theta$ & $\Delta$ & $p$ \\
\midrule
8  & 0.997 & 0.880 & $+0.118$ & $0.003$ \\
12 & 0.997 & 0.822 & $+0.175$ & $<0.001$ \\
16 & 0.975 & 0.810 & $+0.165$ & $0.010$ \\
20 & 0.942 & 0.860 & $+0.082$ & $<0.001$ \\
24 & 0.896 & 0.808 & $+0.087$ & $<0.001$ \\
30 & 0.872 & 0.806 & $+0.066$ & $<0.001$ \\
36 & 0.822 & 0.761 & $+0.061$ & $0.014$ \\
\bottomrule
\end{tabular}
\end{table}

\subsection{Constraint Graph Propagation}
\label{app:zl_propagation}

We test whether a committed cell affects predictions at constraint-linked cells more than unrelated cells. 
For each fill location, we compare a correct fill against a plausible wrong fill and measure the change in top-1 accuracy at other cells, grouped by graph distance from the perturbed cell.

\begin{figure}[h]
\centering
\includegraphics[width=0.5\linewidth]{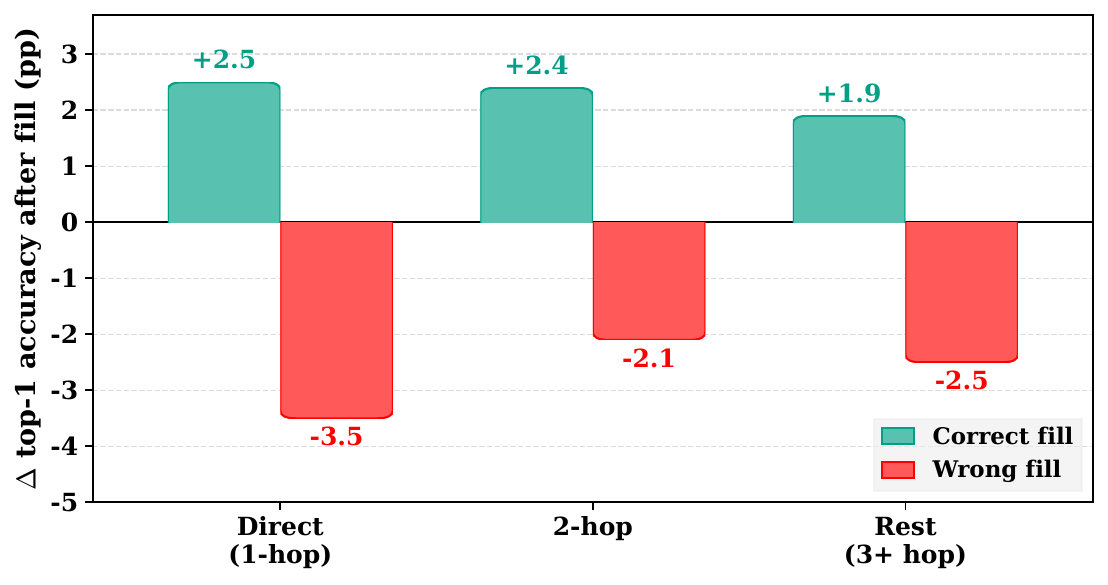}
\caption{Constraint propagation is graph-localized on ZebraLogic-Hard: the correct--wrong response gap is largest at direct clue links and decays with distance.}
\label{fig:propagation}
\end{figure}

Figure~\ref{fig:propagation} shows that the correct--wrong response gap is largest at direct clue links and decays with graph distance. 
This indicates that the model's predictive changes are localized along the puzzle constraint graph.

\subsection{Sparse Correction Recovery on Greedy Failures}
\label{app:zl_sparse_recovery}

\begin{figure}[h]
\begin{minipage}[t]{0.40\linewidth}
\vspace{0pt}
\centering
\captionof{table}{Recovery of greedy failures by ground-truth revelation. Denominator is the $108$ ZebraLogic-Hard puzzles failed by greedy decoding.}
\label{tab:oracle_recovery}
\small
\begin{tabular}{@{}lcc@{}}
\toprule
GT revealed & Solved & Recovery rate \\
\midrule
$0\%$  & $0/108$    & $0.0\%$ \\
$5\%$  & $35/108$   & $32.4\%$ \\
$15\%$ & $94/108$   & $87.0\%$ \\
$25\%$ & $108/108$  & $\mathbf{100.0\%}$ \\
\bottomrule
\end{tabular}
\end{minipage}%
\hfill
\begin{minipage}[t]{0.58\linewidth}
\vspace{0pt}
\centering
\includegraphics[width=\linewidth]{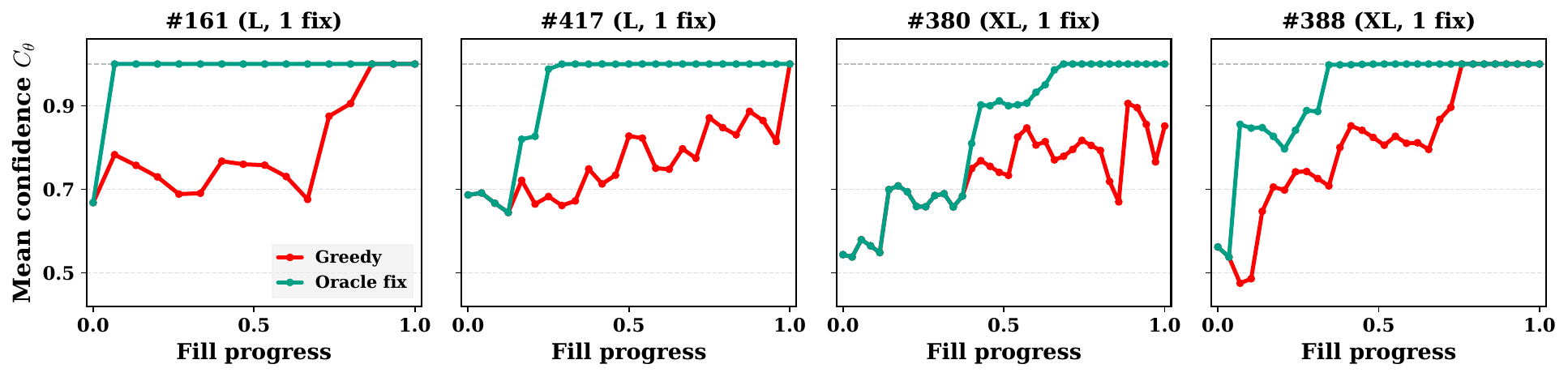}
\caption{Per-puzzle recovery trajectories under sparse ground-truth correction (four representative greedy failures). One or two corrections are typically sufficient to redirect $\mathcal{C}_\theta$ back to convergence.}
\label{fig:oracle_trajectory}
\end{minipage}
\end{figure}

As an oracle diagnostic, we reveal a small fraction of ground-truth cells on the $108$ ZebraLogic-Hard puzzles failed by greedy decoding and rerun greedy inference. 
Table~\ref{tab:oracle_recovery} shows that sparse revelation recovers most failures: $15\%$ revealed cells solve $87.0\%$ of failed puzzles, and $25\%$ solves all failures.

Most recovered puzzles require only one or two revealed cells, suggesting that many greedy failures are locally correctable rather than globally unrecoverable.

\subsection{Autoregressive Inference-Time Refinement Diagnostic}

We evaluate stronger autoregressive inference procedures on ZebraLogic-Hard using the same fine-tuned LLaMA-3.1-8B checkpoint as in the main comparison.
This diagnostic asks whether additional autoregressive search or explicit revision can recover gains comparable to Blackboard inference.
We do not introduce revision-specific supervision, since doing so would alter the matched downstream training setting.
Instead, the perfect-verifier condition provides exact violated-clue feedback while withholding the correct assignments, giving the autoregressive model a favorable inference-time test of whether it can repair a known inconsistency.
ZebraLogic-Hard provides a particularly clean setting for this diagnostic because violated clues can be identified exactly by a programmatic checker.
All compute measurements use the same NVIDIA RTX PRO 6000 GPU.
Wall-clock time, peak memory, and FLOPs are reported per instance for each complete inference procedure.

\subsubsection{Refinement Protocol}
\label{app:zl_ar_refinement_protocol}

Each refinement procedure begins from the same deterministic autoregressive solution used in the main LLaMA greedy baseline.
The resulting Markdown table is serialized into the next prompt as a previous attempt.
The model then generates a complete replacement table rather than editing tokens in place.

\paragraph{Self-critique.}
The model first inspects its current candidate and identifies the clues it believes are violated:

\begin{promptbox}
### SYSTEM:
You are a careful logic-puzzle checker. Given a puzzle and a proposed
solution table, identify which clues the solution VIOLATES and why. If fully
consistent, say ``NO VIOLATIONS''. Do NOT rewrite the table.

### USER:
PUZZLE:
<puzzle text>

PROPOSED SOLUTION:
<current candidate table>

Which clues does this solution violate?
\end{promptbox}

Unless the model returns \texttt{NO VIOLATIONS}, its critique is supplied to the subsequent revision prompt.
This condition therefore tests the full self-diagnosis-and-repair loop using the same fine-tuned checkpoint.

\paragraph{Perfect-verifier refinement.}
A programmatic checker instead returns the exact clues violated by the current candidate.
It does not reveal the correct assignment, any correct cell value, or an edit that would repair a violation.
This removes error localization as a bottleneck while retaining the model's need to infer how a known inconsistency should be repaired into a globally consistent solution.

\paragraph{Revision.}
For either feedback source, we augment the original SFT prompt immediately before its final-answer marker:

\begin{promptbox}
### PREVIOUS ATTEMPT:
<current candidate table>

### FEEDBACK:
<self-generated critique or verifier feedback>

### CORRECTED FINAL SOLUTION:
\end{promptbox}

All generations use deterministic decoding.
We allow at most three refinement rounds and terminate early when the feedback source returns \texttt{NO VIOLATIONS}.
A malformed or infeasible regenerated table is counted as incorrect.

\subsubsection{Tiered Refinement Results}

\begin{table}[h]
\centering
\caption{Autoregressive refinement results by ZebraLogic-Hard tier.
All refinement methods start from the same deterministic LLaMA-3.1-8B solution used in the main greedy baseline and are evaluated on the full $n=500$ test set.}
\label{tab:zl_ar_refinement_by_tier}
\small
\begin{tabular}{@{}lccccc@{}}
\toprule
Inference & S & M & L & XL & All \\
\midrule
Greedy initial solution
& 80.8 & 48.0 & 8.8 & 0.8 & 34.6 \\
Self-critique (up to 3 rounds)
& 68.0 & 33.6 & 3.2 & 0.0 & 26.2 \\
Perfect-verifier refinement (up to 3 rounds)
& 82.4 & 51.2 & 11.2 & 1.6 & 36.6 \\
\bottomrule
\end{tabular}
\end{table}

Self-critique decreases accuracy across every difficulty tier, indicating that self-generated diagnosis can destabilize candidate solutions rather than reliably repair them.
Perfect-verifier feedback avoids this degradation and improves overall accuracy from $34.6\%$ to $36.6\%$, but the gain remains modest even when the exact violated clues are supplied.
In particular, performance remains low on the Large and X-Large tiers despite exact violated-clue feedback.

\subsubsection{Hardware-Matched Compute}
\label{app:zl_ar_hardware_compute}

\begin{table}[h]
\centering
\caption{Hardware-matched autoregressive inference-time diagnostic on ZebraLogic-Hard.
All methods use the same fine-tuned LLaMA-3.1-8B checkpoint and are measured on an NVIDIA RTX PRO 6000 GPU.}
\label{tab:zl_ar_compute}
\small
\setlength{\tabcolsep}{6pt}
\begin{tabular}{@{}lrrrr@{}}
\toprule
Inference & Time / inst. & Peak memory & FLOPs / inst. & Accuracy (\%) \\
\midrule
Greedy                      & 2.40\,s  & 17.1\,GB & 8.2\,T  & 34.6 \\
Beam search ($B=8$)         & 2.61\,s  & 18.0\,GB & 17.2\,T & 34.8 \\
Beam search ($B=64$)        & 5.59\,s  & 25.7\,GB & 85.1\,T & 35.0 \\
\midrule
Self-critique               & 19.14\,s & 17.1\,GB & 59.8\,T & 26.2 \\
Perfect-verifier refinement & 6.03\,s  & 17.1\,GB & 20.5\,T & 36.6 \\
\bottomrule
\end{tabular}
\end{table}

Widening beam search from $B=8$ to $B=64$ yields only a $0.2$\,pp accuracy gain while increasing FLOPs by $4.9{\times}$.
Self-critique reduces accuracy by $8.4$\,pp relative to greedy decoding, while perfect-verifier refinement improves it by only $2.0$\,pp despite exact violated-clue feedback.
Thus, on ZebraLogic-Hard, substantially stronger autoregressive inference---including wider search, explicit revision, and oracle-assisted error localization---does not recover gains comparable to Blackboard inference.
This is a ZebraLogic-specific diagnostic: autoregressive search can have task-dependent effects, as illustrated by the substantially stronger beam-search result on JSSP.

\subsection{dLLM Baselines and Inference Ablations}
\label{app:zl_ablations_baselines}

We describe the dLLM baselines used in Section~\ref{sec:mdm_baseline} and report inference-time ablations on ZebraLogic-Hard. 
Unless stated otherwise, all methods use the same SFT-trained LLaDA-8B backbone.

\subsubsection{PRISM}
\label{app:prism_baseline}

We implement PRISM following~\citet{kim2025fine} by adding a remasking head to the SFT-trained LLaDA-8B backbone:
$\mathrm{LayerNorm} \to \mathrm{Linear}(d{\to}d) \to \mathrm{GELU} \to \mathrm{Linear}(d{\to}1)$ with $d=4096$.
The head is trained jointly with the unmasking objective using
\[
\mathcal{L}
=
\mathcal{L}_{\mathrm{unmask}}
+
\lambda
\frac{1}{|\mathcal{R}|}
\sum_{i\in\mathcal{R}}
\big(\mathrm{softplus}(h_i)-y_i\big)^2,
\quad
y_i=(1-\alpha_t)\mathbf{1}[\tilde{x}_i\neq x_i]+\alpha_t v_i,
\quad
\alpha_t=(1-t)^2,
\]
where $v_i$ is the constraint-violation count from the puzzle generator.
All other LoRA and SFT settings match Appendix~\ref{app:sft_recipe}.

At inference time, a clean token at position $i$ is remasked when its raw head output $h_i$ exceeds a threshold $\mathit{thr}$; unmasking and remasking are interleaved for up to $150$ steps.
We swept $\mathit{thr} \in \{0.7,0.8,0.9,0.95\}$ and sampling temperature $T \in \{0.0,0.3,0.7,1.0\}$, and report the best observed configuration, $\mathit{thr}=0.8$ and $T=1.0$, which reaches $79.2\%$ on ZebraLogic-Hard.

\subsubsection{Blackboard Search and Trigger Ablations}
\label{app:zl_invocation}

We additionally consider \emph{Blackboard Search}, a controlled variant that isolates Blackboard's mean-confidence-guided search component.
Unlike full Blackboard, it applies depth-$3$ search at fixed intervals without confidence-drop-based within-trajectory control or backtracking.
Every $3$ committed cells, each masked position is expanded over its top-$5$ candidate values, rolled out greedily for depth $3$, and scored by the projected final $\mathcal{C}_\theta$.
The best length-$3$ path is committed.

We then ablate puzzle-level triggering for both intervention types.
Triggered settings invoke the corresponding intervention only when
\[
\min_{i \geq 0.8N}\mathcal{C}_{\theta}(\rvx_{t_i}) < 1.0.
\]

\begin{table}[h]
\centering
\caption{Trigger and intervention ablation on ZebraLogic-Hard. The bottom two rows use the validation-selected trigger $(\rho,\tau)=(0.8,1.0)$.}
\label{tab:zl_invocation}
\small
\begin{tabular}{@{}l cc@{}}
\toprule
Setting & NFE & Accuracy \\
\midrule
Greedy                          & 35  & 78.4\% \\
\midrule
Blackboard Search (w/o trigger) & 866 & 84.6\% \\
Blackboard (w/o trigger)        & 280 & 86.8\% \\
\midrule
Blackboard Search               & 273 & 90.2\% \\
\rowcolor{xslategray!8}
\textbf{Blackboard}             & \textbf{129} & \textbf{90.4\%} \\
\bottomrule
\end{tabular}
\end{table}

Puzzle-level triggering improves both intervention types while reducing inference effort.
For Blackboard Search, triggering improves accuracy from $84.6\%$ to $90.2\%$ while reducing average NFE from $866$ to $273$.
For full Blackboard, triggering improves accuracy from $86.8\%$ to $90.4\%$ while reducing average NFE from $280$ to $129$.
Among the triggered variants, Blackboard achieves essentially the same accuracy as Blackboard Search ($90.4\%$ versus $90.2\%$) while using less than half the inference effort ($129$ versus $273$ NFE).

\subsubsection{Lookahead Depth}
\label{app:zl_lookahead}

We vary Blackboard lookahead depth while holding the validation-selected trigger
$(\rho,\tau)=(0.8,1.0)$, branching factor $k=5$, and anchor threshold
$\theta_{\mathrm{anchor}}=0.90$ fixed.

\begin{table}[h]
\centering
\caption{Lookahead-depth ablation for Blackboard on ZebraLogic-Hard.
Each cell reports solve rate (\%).}
\label{tab:la_ablation}
\small
\begin{tabular}{@{}lccccc@{}}
\toprule
Depth $d$ & S & M & L & XL & All \\
\midrule
\rowcolor{xslategray!8}
\textbf{$d{=}3$} & \textbf{100.0} & \textbf{99.2} & \textbf{87.2}
& 75.2 & \textbf{90.4} \\
$d{=}4$ & \textbf{100.0} & 97.6 & 84.0 & \textbf{76.8} & 89.6 \\
$d{=}5$ & \textbf{100.0} & 97.6 & 84.8 & 75.2 & 89.4 \\
\bottomrule
\end{tabular}
\end{table}

Depth $d=3$ performs best overall. Increasing lookahead depth provides no
consistent benefit: $d=4$ slightly improves the X-Large tier but reduces
performance on Medium and Large puzzles, while $d=5$ is lower overall.

\subsection{Official ZebraLogic Benchmark}
\label{app:zl_official}

In the main text, we evaluate on ZebraLogic-Hard, a $500$-instance variant constructed to stress-test partial-assignment feasibility. For completeness, we also report results on the official ZebraLogic benchmark of \citet{lin2025zebralogic} ($n=1000$ puzzles).

\paragraph{Adaptation of the official benchmark.}
We adapt the official ZebraLogic puzzles to our single-token grid-filling format while preserving their constraint structure. 
Each clue is parsed into the canonical predicate vocabulary used for ZebraLogic-Hard, and category/value names are replaced with single-token aliases. 
The conversion succeeds on all $1000$ puzzles, and all converted constraints remain solvable by Z3 with the original solution.

\paragraph{Setup.}
All methods use the same SFT-trained LLaDA-8B checkpoint as in the main text
and the same Blackboard inference framework with $d{=}3$, $k{=}5$, and
$\theta_{\mathrm{anchor}}{=}0.90$. For this evaluation, Blackboard uses
$(\rho,\tau)=(2/3,1.0)$ and triggers when
\[
\min_{i \geq \rho N}\mathcal{C}_{\theta}(\rvx_{t_i}) < \tau.
\]

\begin{table}[h]
\centering
\caption{Results on the adapted official ZebraLogic benchmark ($n=1000$,
search-space bins Small/Medium/Large/X-Large following
\citet{lin2025zebralogic}). Blackboard uses $(\rho,\tau)=(2/3,1.0)$;
non-triggered puzzles keep their greedy result.}
\label{tab:zl_official}
\small
\setlength{\tabcolsep}{6pt}
\begin{tabular}{@{}l cccccc@{}}
\toprule
Model  & NFE & S & M & L & XL & All \\
\midrule
LLaDA-8B (greedy)                  & 32       & 96.6 & 89.6 & 74.5 & 50.0 & 80.9 \\
LLaDA-8B (Blackboard, w/o trigger) & 223      & 96.6 & 93.6 & 78.0 & 48.5 & 82.4 \\
\rowcolor{xslategray!8}
\textbf{LLaDA-8B (Blackboard)}     & \textbf{145} & \textbf{96.6} & \textbf{93.9} & \textbf{83.5} & \textbf{58.5} & \textbf{85.6} \\
\bottomrule
\end{tabular}
\end{table}

\paragraph{Result.}
Blackboard improves over greedy by $+4.7$\,pp ($80.9\%\to85.6\%$), with
larger gains on Large and X-Large puzzles.
It also outperforms its untriggered variant while using fewer NFEs
($145$ versus $223$), providing supplementary evidence that
confidence-triggered correction remains useful after conversion of the
official benchmark to the single-token grid interface.

% ============================================================
% NURSE ROSTERING ADDITIONAL RESULTS
% ============================================================
\section{Nurse Rostering Additional Results}
\label{app:nr_additional}

We provide additional Nurse Rostering analyses: confidence separation between solved and failed trajectories, performance across conflict-based difficulty bands, and the effect of confidence-triggered steering.

\subsection{Confidence Separation by Difficulty}
\label{app:nr_confidence_by_difficulty}

We partition the $500$ held-out instances into four equal-size difficulty bands using the number of conflicts encountered by the Z3 solver during constraint solving: d0 contains instances with $0$--$1$ conflicts, d1 with $2$--$4$, d2 with $5$--$9$, and d3 with $10$--$31$ conflicts.

\begin{figure}[h]
    \centering
    \vspace{-0.10in}
    \begin{minipage}[t]{0.45\linewidth}
        \centering
        \includegraphics[width=\linewidth]{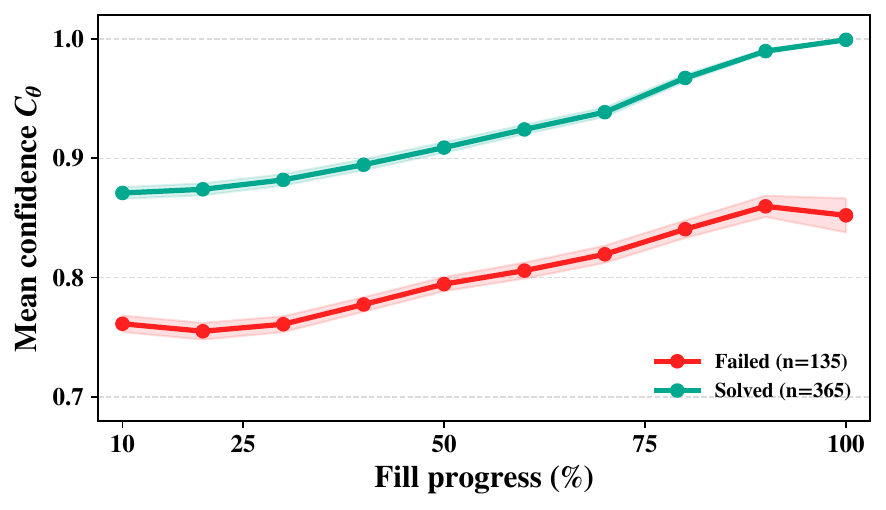}
    \end{minipage}
    \hfill
    \begin{minipage}[t]{0.53\linewidth}
        \centering
        \includegraphics[width=\linewidth]{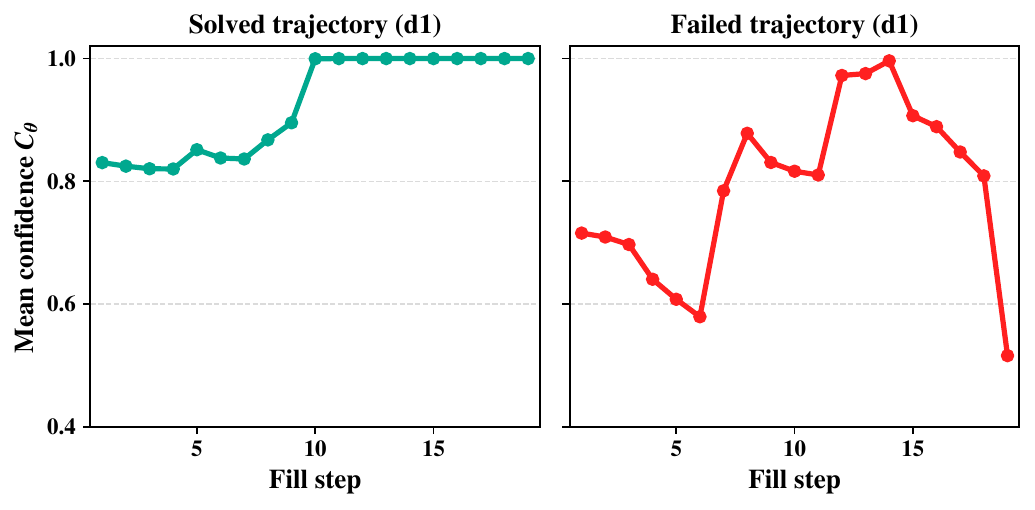}
    \end{minipage}
    \vspace{-0.08in}
    \caption{\textbf{Mean confidence trajectories on Nurse Rostering.}
    (\textbf{Left}) Aggregated over evaluation instances, solved rosters exhibit
    higher mean confidence than failed ones.
    (\textbf{Right}) Illustrative d1 trajectories: the solved trajectory converges
    toward high confidence, whereas the shown failed trajectory undergoes substantial
    late-stage variation.}
    \label{fig:nr_mc}
    \vspace{-0.10in}
\end{figure}

Figure~\ref{fig:nr_mc} reproduces the aggregate solved--failed confidence separation observed on ZebraLogic. The right panel provides qualitative trajectory examples, while the quantitative analysis below measures this separation over all held-out instances and across difficulty bands.

\begin{table}[h]
\centering
\caption{Separation between solved and failed Nurse Rostering trajectories,
measured by Cohen's $d$ of trajectory-averaged mean confidence.}
\label{tab:nr_confidence_by_difficulty}
\small
\begin{tabular}{@{}lccccc@{}}
\toprule
Conflict band & d0 & d1 & d2 & d3 & All \\
\midrule
Cohen's $d$ & 2.48 & 2.18 & 2.03 & 1.82 & 2.16 \\
\bottomrule
\end{tabular}
\end{table}

Mean confidence separates solved from failed trajectories in every difficulty band. 
Although the separation decreases moderately with conflict density, it remains large even on d3, supporting mean confidence as a reliability signal for selective correction.

\subsection{Performance Across Conflict Bands}
\label{app:nr_by_difficulty}

Table~\ref{tab:nr_main} reports exact feasibility across the Z3-conflict difficulty bands defined above, together with the full comparison to frontier and same-scale autoregressive baselines.

\begin{table}[t]
\centering
\caption{Evaluation results on Nurse Rostering.
Instances are grouped by Z3-conflict difficulty bands (d0--d3);
\textsc{All} is exact feasibility over the full evaluation set.}
\label{tab:nr_main}

\fontsize{9.0pt}{10.0pt}\selectfont
\setlength{\tabcolsep}{7.5pt}
\renewcommand{\arraystretch}{0.87}
\setlength{\heavyrulewidth}{0.45pt}
\setlength{\lightrulewidth}{0.30pt}
\setlength{\cmidrulewidth}{0.30pt}

\begin{tabular}{@{}lllrccccc@{}}
\toprule
\textbf{Class}
& \textbf{Model}
& \textbf{Inference}
& \textbf{NFE}
& \textbf{d0}
& \textbf{d1}
& \textbf{d2}
& \textbf{d3}
& \textbf{All} \\
\midrule

\multirow{4}{*}{
\coloredhl{xxpurple}{\textbf{Frontier LLMs}}
}
& \multicolumn{2}{l}{GPT-5.4}
& --- & 52.0 & 8.0 & 3.2 & 0.0 & 15.8 \\

& \multicolumn{2}{l}{GPT-5.4 (CoT prompting)}
& --- & 91.2 & 67.2 & 31.2 & 14.4 & 51.0 \\

& \multicolumn{2}{l}{GPT-5.4 Thinking}
& --- & 92.0 & 90.4 & 84.0 & 77.6 & 86.0 \\

& \multicolumn{2}{l}{Gemini 2.5 Pro}
& --- & 73.6 & 41.6 & 14.4 & 8.0 & 34.4 \\

\midrule

\multirow{3}{*}{
\coloredhl{xxpurple}{\textbf{8B-LLM}}
}
& LLaMA-3.1-8B & Greedy
& 97 & 39.2 & 8.0 & 5.6 & 2.4 & 13.8 \\

& LLaMA-3.1-8B & Beam search
& 773 & 42.4 & 8.0 & 6.4 & 2.4 & 14.8 \\

& LLaMA-3.1-8B & Sample-BoN
& 773 & 40.8 & 8.0 & 4.8 & 2.4 & 14.0 \\

\midrule

\multirow{2}{*}{
\coloredhl{xxpurple}{\textbf{8B-dLLM}}
}
& LLaDA-8B & Greedy
& 22
& 95.2
& 84.0
& 60.0
& 52.8
& 73.0 \\

& \cellcolor{xxpurple!20}\textbf{LLaDA-8B}
& \cellcolor{xxpurple!20}\textbf{Blackboard}
& \cellcolor{xxpurple!20}\textbf{120}
& \cellcolor{xxpurple!20}\textbf{95.2}
& \cellcolor{xxpurple!20}\textbf{85.6}
& \cellcolor{xxpurple!20}\textbf{65.6}
& \cellcolor{xxpurple!20}\textbf{59.2}
& \cellcolor{xxpurple!20}\textbf{76.4} \\

\bottomrule
\end{tabular}
\end{table}

Performance generally declines as constraint interactions increase.
Among the same-scale models, Blackboard improves over LLaDA greedy in d1--d3, with gains increasing on the more conflict-dense bands: $+1.6$\,pp on d1, $+5.6$\,pp on d2, and $+6.4$\,pp on d3.
The d0 rate is unchanged at $95.2\%$, leaving little headroom for correction.
GPT-5.4 Thinking provides the strongest frontier result at $86.0\%$ overall, illustrating that the relative advantage of Blackboard is task-dependent.

\subsection{Why Confidence Triggering Matters}
\label{app:nr_trigger_compute}

The corrective cascade is not applied to every roster.  
The validation-selected trigger uses the mean confidence over the final $10\%$ of filling steps and fires when this statistic falls below $\tau=0.95$, i.e., $(\rho,\tau)=(0.90,0.95)$.
Table~\ref{tab:nr_trigger_diagnostic} shows that always applying the cascade is harmful: it can overwrite already-correct rosters. 
The confidence trigger protects high-confidence trajectories while concentrating correction on the subset with elevated failure risk.

\begin{table}[h]
\centering
\caption{Effect of confidence-triggered steering on Nurse Rostering.
The full cascade is evaluated on all $500$ instances; Blackboard applies it
only to triggered instances and otherwise retains the greedy roster.}
\label{tab:nr_trigger_diagnostic}
\small
\begin{tabular}{@{}lccc@{}}
\toprule
Method & Cascade applied & Mean NFE & Exact feasibility (\%) \\
\midrule
LLaDA-8B greedy & -- & 22 & 73.0 \\
Cascade on every instance & 100\% & 325 & 57.4 \\
\rowcolor{xslategray!8}
\textbf{Confidence-triggered Blackboard}
& \textbf{19.6\%} & \textbf{120} & \textbf{76.4} \\
\bottomrule
\end{tabular}
\end{table}

The trigger fires on $98/500$ instances.  Among these triggered instances,
greedy decoding solves $4$, whereas the corrective cascade yields $21$
correct rosters.  This corresponds to $19$ failure-to-solution changes and
$2$ regressions.  The remaining $402$ non-triggered instances retain their
greedy output, including $361$ already-correct rosters.  Thus, the trigger is
necessary both for accuracy and for limiting corrective compute.

% ============================================================
% JSSP ADDITIONAL RESULTS
% ============================================================
\section{JSSP Additional Results}
\label{app:jssp_additional}

\subsection{Main Evaluation Results}
\label{app:jssp_main_results}

Table~\ref{tab:jssp_main} reports the full comparison on JSSP, including frontier LLMs and matched 8B autoregressive and diffusion models.
The main text summarizes the headline comparison, while the analyses below examine performance by problem size and the role of confidence-triggered computation.

\begin{table}[h]
\centering
\caption{Evaluation results on JSSP.}
\label{tab:jssp_main}

\fontsize{9.0pt}{10.0pt}\selectfont
\setlength{\tabcolsep}{12.5pt}
\renewcommand{\arraystretch}{0.87}
\setlength{\heavyrulewidth}{0.45pt}
\setlength{\lightrulewidth}{0.30pt}
\setlength{\cmidrulewidth}{0.30pt}

\begin{tabular}{@{}lllrcc@{}}
\toprule
\textbf{Class}
& \textbf{Model}
& \textbf{Inference}
& \textbf{NFE}
& \textbf{\% opt}
& \textbf{\% within 5\%} \\
\midrule

\multirow{4}{*}{
\coloredhl{xxpurple}{\textbf{Frontier LLMs}}
}
& \multicolumn{2}{l}{GPT-5.4}
& --- & 11.0 & 16.5 \\

& \multicolumn{2}{l}{GPT-5.4 (CoT prompting)}
& --- & 30.0 & 37.2 \\

& \multicolumn{2}{l}{GPT-5.4 Thinking}
& --- & 77.0 & 90.2 \\

& \multicolumn{2}{l}{Gemini 2.5 Pro}
& --- & 27.8 & 35.0 \\

\midrule

\multirow{3}{*}{
\coloredhl{xxpurple}{\textbf{8B-LLM}}
}
& LLaMA-3.1-8B & Greedy
& 85 & 45.0 & 66.2 \\

& LLaMA-3.1-8B & Beam search
& 850 & 71.2 & 89.5 \\

& LLaMA-3.1-8B & Sample-BoN
& 850 & 63.0 & 83.8 \\

\midrule

\multirow{2}{*}{
\coloredhl{xxpurple}{\textbf{8B-dLLM}}
}
& LLaDA-8B & Greedy
& 22 & 70.0 & 91.0 \\

& \cellcolor{xxpurple!20}\textbf{LLaDA-8B}
& \cellcolor{xxpurple!20}\textbf{Blackboard}
& \cellcolor{xxpurple!20}\textbf{149}
& \cellcolor{xxpurple!20}\textbf{80.2}
& \cellcolor{xxpurple!20}\textbf{95.2} \\

\bottomrule
\end{tabular}
\end{table}
 
\subsection{Per-Size Results}
\label{app:jssp_by_size}
 
Table~\ref{tab:jssp_by_size} reports JSSP results by grid size. 
For each size, \% optimal and \% within $5\%$ use all instances at that size as the denominator, with invalid outputs counted as failures; mean $\mathrm{ms\_ratio}$ is computed over valid outputs only.
 
\begin{table}[h]
\centering
\caption{JSSP per-size results. Each cell reports \% optimal\,/\,\% within
$5\%$\,/\,mean $\mathrm{ms\_ratio}$. Mean $\mathrm{ms\_ratio}$ is computed
over valid outputs only. Blackboard uses the validation-selected setting
$(\rho,\tau)=(0.5,0.7)$ with $N{=}10$
(Section~\ref{sec:algorithm}). The bottom row corresponds to the Blackboard
entry in Table~\ref{tab:jssp_main}.}
\label{tab:jssp_by_size}
\scriptsize
\setlength{\tabcolsep}{2.5pt}
\begin{tabular}{@{}l ccccccc@{}}
\toprule
Method & 3$\times$3 & 4$\times$3 & 4$\times$4 & 5$\times$4 & 5$\times$5 & 6$\times$6 & 8$\times$8 \\
\midrule
GPT-5.4                             & 35/35/1.11 & 27/39/1.11 & 17/23/1.20 &  2/10/1.23 &  1/ 6/1.29 &  0/ 0/1.37 &  0/ 0/1.58 \\
GPT-5.4 (CoT)                       & 82/82/1.02 & 77/84/1.07 & 47/60/1.06 &  7/21/1.17 &  6/ 9/1.19 &  0/ 2/1.36 &  0/ 0/2.19 \\
GPT-5.4 (thinking)                 &100/100/1.00& 96/100/1.00& 94/99/1.00 & 79/100/1.01& 73/96/1.01 & 40/60/1.05 &  0/ 8/1.15 \\
Gemini 2.5 Pro                & 88/88/1.01 & 46/57/1.10 & 43/59/1.09 & 15/22/1.19 &  7/15/1.28 &  2/ 2/1.53 &  0/ 0/1.59 \\
\midrule
LLaMA-3.1-8B (beam-10)              &100/100/1.00& 98/100/1.00& 90/97/1.00 & 72/92/1.01 & 67/91/1.01 & 22/78/1.04 &  0/ 0/11.12 \\
LLaMA-3.1-8B (sample-BoN $T{=}1.0$) & 90/90/1.01 & 95/98/1.00 & 81/91/1.01 & 58/88/1.02 & 57/85/1.02 & 21/67/1.04 &  0/ 0/12.19 \\
\midrule
LLaDA-8B (greedy)                                  & 98/98/1.00 & 86/100/1.01& 91/97/1.00 & 69/95/1.01 & 66/90/1.01 & 33/84/1.03 & 0/ 8/1.11 \\
LLaDA-8B (BoN, w/o trigger)                    &100/100/1.00& 93/100/1.00& 97/99/1.00 & 86/98/1.01 & 79/99/1.01 & 60/97/1.01 & 8/31/1.07 \\
\rowcolor{xslategray!8}
\textbf{LLaDA-8B (Blackboard)}
& 98/98/1.00 & 91/100/1.00
& 96/97/1.00 & 83/96/1.01
& 78/99/1.01 & 53/97/1.01
& 8/31/1.07 \\
\bottomrule
\end{tabular}
\end{table}
 
Blackboard improves most on the larger grid sizes, where greedy decoding is
more often suboptimal. On the hardest $8{\times}8$ tier, it attains the
joint-highest within-$5\%$ rate in the table.

\subsection{Blackboard vs.\ Untriggered BoN}
\label{app:jssp_trigger_compute}
 
Table~\ref{tab:jssp_trigger_compute} compares Blackboard with untriggered BoN on JSSP.
 
\begin{table}[h]
\centering
\caption{Blackboard versus untriggered BoN on JSSP. Blackboard uses the
validation-selected setting $(\rho,\tau)=(0.5,0.7)$.}
\label{tab:jssp_trigger_compute}
\small
\begin{tabular}{@{}l ccc@{}}
\toprule
Method & Mean NFE & \% within 5\% & Mean ms ratio \\
\midrule
dLLM greedy                              & 22  & 91.0 & 1.013 \\
dLLM BoN (w/o trigger)                   & 217 & 96.2 & 1.007 \\
\rowcolor{xslategray!8}
\textbf{dLLM Blackboard}                 & \textbf{149} & \textbf{95.2} & \textbf{1.008} \\
\bottomrule
\end{tabular}
\end{table}
 
Blackboard reduces average NFEs by $31\%$ relative to untriggered BoN, with a
$1.0$\,pp drop in within-$5\%$ rate.
 
\subsection{Per-Trajectory Quality Before BoN Selection}
\label{app:jssp_per_trajectory}

We compare individual stochastic trajectories before Best-of-$N$ selection on the same JSSP puzzles where the Blackboard trigger fires.
Both LLaMA-3.1-8B and LLaDA-8B are sampled with $N=10$ and $T=1.0$; metrics are computed over valid trajectories.

\begin{table}[h]
\centering
\caption{Per-trajectory JSSP quality before Best-of-$N$ selection on the
$226$ puzzles for which the validation-selected Blackboard trigger fires.
Both models are sampled with $N=10$ and $T=1.0$; metrics are computed over
valid trajectories. Unique makespans is the average number of distinct
makespans among the $N=10$ samples per puzzle.}
\label{tab:jssp_per_trajectory}
\small
\begin{tabular}{@{}lcccc@{}}
\toprule
Model & Valid trajectories & \% optimal & Median $\mathrm{ms\_ratio}$ & Unique makespans \\
\midrule
LLaMA-3.1-8B & 2{,}164 & 35.1 & 1.039 & 2.65 \\
\rowcolor{xslategray!8}
\textbf{LLaDA-8B} & \textbf{2{,}219} & \textbf{51.0} & \textbf{1.000} & \textbf{2.88} \\
\bottomrule
\end{tabular}
\end{table}

LLaDA produces higher-quality individual trajectories while retaining
comparable schedule diversity, suggesting that the BoN gain is not explained
solely by broader sampling diversity.

\section{Validation Selection, Sensitivity, and Uncertainty}
\label{app:threshold}

\subsection{Confidence Trigger Selection}
\label{app:trigger_selection}

We select trigger hyperparameters before test evaluation using held-out training
trajectories that are disjoint from the reported test sets.
For each trajectory, a task-specific greedy failure is treated as the positive class and trigger firing as the prediction. 
We use a task-specific late-phase confidence statistic, then evaluate candidate late-phase fractions $\rho$ and thresholds $\tau$ and freeze the pair maximizing a task-appropriate failure-detection score.

The task-specific late-phase confidence statistic referenced in Section~\ref{sec:algorithm} is defined as follows. 
For ZebraLogic and JSSP,
we use the minimum mean confidence within the late phase:
\[
s_\rho(x) =
\min_{i \in \{\lceil \rho N\rceil,\ldots,N-1\}}
\mathcal{C}_{\theta}(x_{t_i}).
\]
For Nurse Rostering, we use the late-phase mean confidence to reduce
sensitivity to isolated confidence dips:
\[
s_\rho^{\mathrm{NR}}(x) =
\frac{1}{N-\lceil\rho N\rceil}
\sum_{i=\lceil\rho N\rceil}^{N-1}
\mathcal{C}_{\theta}(x_{t_i}).
\]
In all settings, additional inference is invoked when the corresponding
late-phase confidence statistic falls below $\tau$.

A late-phase window avoids two degenerate choices.
Early in inference, most cells remain masked, and solved and failed trajectories are less separated;
near completion, few masked cells remain and confidence can saturate.
We therefore select $\rho$ jointly with $\tau$, rather than fixing either value by hand.

\begin{table}[h]
\centering
\caption{Pre-test trigger selection from greedy trajectories. Precision and
recall refer to detecting greedy failures. ZebraLogic and JSSP select the pair
with maximum $F_1$; Nurse Rostering selects by $F_{0.5}$ to prioritize
precision and avoid overwriting already-correct rosters.}
\label{tab:trigger_selection}
\small
\setlength{\tabcolsep}{5pt}
\begin{tabular}{@{}l l c ccc@{}}
\toprule
Task & Statistic & $(\rho,\tau)$
& Precision (\%) & Recall (\%) & Selection score \\
\midrule
ZebraLogic-Hard & late-phase min  & $(0.80,1.00)$ & 92.7 & 85.1 & $F_1=88.7$ \\
Nurse Rostering & late-phase mean & $(0.90,0.95)$ & 97.6 & 71.0 & $F_{0.5}=90.8$ \\
JSSP            & late-phase min  & $(0.50,0.70)$ & 21.4 & 79.6 & $F_1=33.8$ \\
\bottomrule
\end{tabular}
\end{table}

The task-specific confidence statistic is fixed before the $\rho$--$\tau$
selection, and the selected settings are frozen before test evaluation and
used for all main-text and task-specific results.
Confidence separates optimal from suboptimal schedules less sharply on JSSP
than it separates exact-feasibility failures on ZebraLogic and Nurse Rostering.
Nevertheless, the selected trigger recalls $79.6\%$ of non-optimal greedy
schedules while selectively allocating objective-ranked BoN computation.

\subsection{Corrective-Operator Sensitivity on ZebraLogic}
\label{app:operator_sensitivity}

We vary the anchor confidence threshold $\alpha$ and candidate width $k$ on the greedy-failure subset while holding depth $d=3$ fixed.
Recovery is unchanged across $\alpha\in\{0.85,0.90,0.95\}$, so Table~\ref{tab:zl_operator_sensitivity} summarizes sensitivity to candidate width.
We use $\alpha=0.90$ and $k=5$ in the main ZebraLogic experiments.

\begin{table}[h]
\centering
\caption{ZebraLogic corrective-operator sensitivity on greedy failures
($n=25$).
Recovery is unchanged across $\alpha\in\{0.85,0.90,0.95\}$; NFE ranges
report variation across these $\alpha$ settings.}
\label{tab:zl_operator_sensitivity}
\small
\begin{tabular}{@{}lccc@{}}
\toprule
Candidate width $k$ & 3 & 5 (main) & 7 \\
\midrule
Recovered instances & 21/25 & 22/25 & 23/25 \\
Recovery rate & 84\% & 88\% & 92\% \\
Mean NFE & 286--317 & 440--484 & 594--635 \\
\bottomrule
\end{tabular}
\end{table}

Performance is stable across the tested anchor thresholds, while increasing
candidate width yields a gradual accuracy--compute trade-off.

\subsection{Evaluation Uncertainty and Paired Comparisons}
\label{app:uncertainty}

We quantify uncertainty for selected headline results.
For the ZebraLogic-Hard greedy comparison, both models are evaluated on the
same $500$ instances, so we use a paired nonparametric bootstrap with
$20{,}000$ resamples for the accuracy gap.
All marginal rates use $95\%$ Wilson score intervals.
These intervals characterize finite held-out evaluation uncertainty rather
than variation across independently retrained checkpoints.

\begin{table}[h]
\centering
\caption{Uncertainty for selected main-task results.
Marginal rates report $95\%$ Wilson intervals; the ZebraLogic-Hard greedy
gap uses a paired bootstrap with $20{,}000$ resamples.}
\label{tab:main_uncertainty}
\small
\setlength{\tabcolsep}{5pt}
\begin{tabular}{@{}llc@{}}
\toprule
Task & Result & Estimate [95\% CI] \\
\midrule
ZebraLogic-Hard
& LLaDA greedy
& 78.4 [74.6, 81.8] \\
& LLaMA greedy
& 34.6 [30.6, 38.9] \\
& Paired greedy gap
& \textbf{+43.8 [+39.2, +48.2]} \\
& LLaDA Blackboard
& 90.4 [87.5, 92.7] \\
\midrule
Nurse Rostering
& LLaDA greedy
& 73.0 [68.9, 76.7] \\
& LLaDA Blackboard
& 76.4 [72.5, 79.9] \\
\midrule
JSSP
& LLaMA beam-10 optimality
& 71.2 [66.6, 75.5] \\
& LLaDA Blackboard optimality
& 80.2 [76.1, 83.9] \\
& LLaDA Blackboard within $5\%$
& 95.2 [92.7, 96.9] \\
\bottomrule
\end{tabular}
\end{table}

On ZebraLogic-Hard, the paired LLaDA--LLaMA greedy gap is positive in every
difficulty tier: $+19.2$, $+46.4$, $+59.2$, and $+50.4$ pp on S, M, L, and XL,
respectively, with the $95\%$ paired-bootstrap interval excluding zero in all
four tiers.
Across the full evaluation set, LLaDA alone solves $224$ instances whereas
LLaMA alone solves $5$; the continuity-corrected McNemar test gives
$\chi^2=207.5$ ($p<10^{-40}$).
The marginal intervals further show that the main cross-domain conclusions
are stable under held-out evaluation uncertainty, while the smaller
Nurse Rostering gain is correspondingly less separated.

% ============================================================
% FRONTIER API REFERENCES
% Black-box reference results for the three main-task settings.
% ============================================================
\section{Frontier API References}
\label{app:frontier_ref}

The main-text results include frontier API models as black-box reference points
on ZebraLogic-Hard, Nurse Rostering, and JSSP, using the same task-specific
input-output format as the fine-tuned models. These results are not
compute-normalized comparisons: model scale, serving configuration, training
data, reasoning budgets, and hidden internal computation are not exposed for
the API models.

For ZebraLogic-Hard and JSSP, we additionally report observed client-side
wall-clock latency and emitted-output-token counts. These are practical
descriptors rather than measurements of comparable inference compute. Nurse
Rostering API results are reported as accuracy references only.

\paragraph{Measurement protocol.}
We use a fixed prompt template per task and mode and do not tune prompts,
sampling settings, or reasoning budgets on a per-puzzle basis. Wall-clock
latency is measured client-side from request submission to complete response
receipt, including network and provider-side latency observed by the client.
Output token counts come from provider \texttt{usage} fields.

\paragraph{ZebraLogic-Hard.}

\begin{table}[h]
\centering
\caption{Frontier API references on ZebraLogic-Hard ($n=500$).}
\label{tab:frontier_zl}
\small
\setlength{\tabcolsep}{8pt}
\begin{tabular}{@{}l c r r r@{}}
\toprule
Model & Mode & Acc & Median latency & Median output tok. \\
\midrule
GPT-5.4         & standard  & $24.4\%$ & $16.2$\,s  & $965$ \\
GPT-5.4         & CoT       & $35.4\%$ & $31.0$\,s  & $2{,}207$ \\
GPT-5.4         & thinking & $73.2\%$ & $72.9$\,s  & $3{,}934$ \\
Gemini 2.5 Pro  & thinking  & $54.4\%$ & $101.4$\,s & $11{,}000$ \\
\bottomrule
\end{tabular}
\end{table}

GPT-5.4 Thinking is the strongest frontier API reference on ZebraLogic-Hard,
reaching $73.2\%$. The matched fine-tuned LLaDA-8B Blackboard system reaches
$90.4\%$; the API latency and token columns should be interpreted only as
practical descriptors rather than compute-normalized quantities.

\paragraph{Nurse Rostering.}

\begin{table}[h]
\centering
\caption{Frontier API references on Nurse Rostering ($n=500$).
The metric is exact-feasibility rate.}
\label{tab:frontier_nr}
\small
\setlength{\tabcolsep}{8pt}
\begin{tabular}{@{}l c r@{}}
\toprule
Model & Mode & Exact feasibility \\
\midrule
GPT-5.4        & standard & $15.8\%$ \\
GPT-5.4        & CoT      & $51.0\%$ \\
GPT-5.4        & thinking & $86.0\%$ \\
Gemini 2.5 Pro & thinking & $34.4\%$ \\
\bottomrule
\end{tabular}
\end{table}

GPT-5.4 Thinking is the strongest tested frontier reference on Nurse Rostering,
reaching $86.0\%$ exact feasibility. The matched fine-tuned LLaDA-8B
Blackboard system reaches $76.4\%$; therefore, Nurse Rostering is presented as
evidence of cross-domain steering transfer rather than a universal
frontier-model superiority claim.

\paragraph{JSSP.}

\begin{table}[h]
\centering
\caption{Frontier API references on JSSP. All rows use the full $n{=}400$ evaluation set. \% optimal and \% within $5\%$ use the full set as denominator, with invalid outputs counted as failures. Mean $\mathrm{ms\_ratio}$ is computed over valid outputs only.}
\label{tab:frontier_jssp}
\small
\setlength{\tabcolsep}{6pt}
\begin{tabular}{@{}l c rrr r r@{}}
\toprule
Model & Mode & \% opt & \% within 5\% & Mean ms ratio & Med lat & Med tok \\
\midrule
GPT-5.4         & standard  & 11.0 & 16.5 & 1.233 & $4$\,s     & $224$ \\
GPT-5.4         & CoT       & 30.0 & 37.2 & 1.185 & $40$\,s    & $2{,}686$ \\
GPT-5.4         & thinking  & 77.0 & 90.2 & 1.015 & $170.6$\,s & $10{,}472$ \\
Gemini 2.5 Pro  & thinking  & 27.8 & 35.0 & 1.201 & $106.5$\,s & $11{,}758$ \\
\bottomrule
\end{tabular}
\end{table}
 
GPT-5.4 thinking is the strongest frontier API reference on JSSP, reaching $77.0\%$ optimal and $90.2\%$ within $5\%$.
For reference, the matched fine-tuned LLaDA-8B system in Table~\ref{tab:jssp_main} reaches $80.2\%$ optimal and $95.2\%$ within $5\%$; the API latency and token columns should be interpreted only as practical descriptors rather than compute-normalized quantities.

The frontier rows are included only for the three main-task settings.
The rotating-shift roster and ATSP experiments in
Appendix~\ref{app:transfer} are supplementary transfer studies evaluated
primarily through matched-model and within-backbone comparisons.

% ============================================================
% SUPPLEMENTARY TRANSFER STUDIES
% ============================================================
\section{Supplementary Transfer Studies}
\label{app:transfer}

Beyond the three main-task settings, we evaluate the same state-level
confidence interface on two supplementary globally constrained domains.
These studies are transfer evidence rather than additional headline
comparisons: they test whether confidence can support task-appropriate
inference when both the constraint structure and output representation differ
from the main tasks.

The studies also illustrate that Blackboard is not a fixed search operator.
For rotating-shift roster completion, confidence is most informative only at
the trajectory level and is used for verifier-free candidate ranking. For
asymmetric TSP, a computable tour-length objective is available, so confidence
instead gates objective-ranked Best-of-$N$ search, as in JSSP.

\begin{table}[h]
\centering
\caption{Summary of supplementary transfer studies.}
\label{tab:supp_transfer_summary}
\small
\begin{tabular}{@{}l l l l@{}}
\toprule
Domain & Problem type & Confidence role & Corrective action \\
\midrule
Rotating-shift roster &
unique feasibility &
trajectory-level ranking &
$\mathcal{C}_\theta$-BoN \\
Asymmetric TSP &
permutation optimization &
compute allocation &
objective-BoN \\
\bottomrule
\end{tabular}
\end{table}

% ------------------------------------------------------------
% ROTATING-SHIFT ROSTER
% ------------------------------------------------------------
\subsection{Rotating-Shift Roster}
\label{app:rr_additional}

\subsubsection{Task and Setup}

We consider a rotating-shift rostering problem whose combinatorial core is
Latin-square completion. The output is an $n \times n$ staff--day grid, where
each cell contains one of $n$ shift types. Every staff member must work each
shift exactly once, and every shift must be covered exactly once on each day.
A set of given assignments pins a unique feasible completion.

We use $n=7$ grids and a held-out evaluation set of $320$ instances, stratified
into four equal-size difficulty bands of $80$ instances each. Difficulty is the
number of backtracks required by an independent minimum-remaining-values solver:
d0 has zero backtracks, d1 has $5$--$19$, d2 has $20$--$49$, and d3 has
$50$--$149$. The model is not given this statistic.

Unlike ZebraLogic, an incorrect local assignment can remain locally plausible
in this domain. On a triggered instance, we therefore sample $K=10$ complete
rosters and select the candidate with the highest mean
$\mathcal{C}_\theta$ over the final five commit steps (\textsc{last5}).
The statistic and trigger threshold are selected on a separate $240$-instance
validation split. No solver, task objective, or feasibility verifier is used
for candidate ranking.

\subsubsection{Difficulty-Controlled Results}

Table~\ref{tab:rr_by_difficulty} reports exact-match feasibility across
solver-backtrack bands. The confidence-triggered Best-of-$N$ procedure improves
performance in every band, with its largest gain on search-hard d3 instances.

\begin{table}[h]
\centering
\caption{Rotating-shift roster exact-match feasibility (\%) by
solver-backtrack difficulty. Each band contains $80$ held-out instances.}
\label{tab:rr_by_difficulty}
\small
\setlength{\tabcolsep}{4pt}
\begin{tabular}{@{}lccccc@{}}
\toprule
Method & All & d0 & d1 & d2 & d3 \\
\midrule
LLaMA-3.1-8B (greedy)        & 8.8  & 13.8 & 5.0  & 5.0  & 11.2 \\
LLaMA-3.1-8B (beam-10)       & 26.2 & 35.0 & 21.2 & 23.8 & 25.0 \\
LLaMA-3.1-8B (Sample-BoN-10) & 17.5 & 21.2 & 15.0 & 13.8 & 20.0 \\
\midrule
LLaDA-8B (greedy) & 95.3 & 98.8 & 98.8 & 92.5 & 91.2 \\
\rowcolor{xslategray!8}
\textbf{LLaDA-8B (Blackboard)}
& \textbf{97.8} & \textbf{98.8} & \textbf{98.8}
& \textbf{95.0} & \textbf{98.8} \\
\bottomrule
\end{tabular}
\end{table}

The aggregate gain is bounded by the strong greedy baseline, but Blackboard
reduces the number of greedy errors from $15$ to $7$ ($53\%$ reduction). On the
hardest d3 band, it improves exact feasibility by $7.6$\,pp. ZebraLogic-style
local lookahead steering is not effective in this domain, indicating that the
appropriate corrective action depends on the confidence landscape.

\subsubsection{Trigger and Candidate-Selection Diagnostic}

The confidence trigger fires on $15/320$ instances ($4.7\%$), precisely the
$15$ instances failed by greedy decoding. Thus, it has $100\%$ precision and
recall for greedy failures on this evaluation set, with no regressions from
intervening on already-correct rosters.

\begin{table}[h]
\centering
\caption{Selective confidence-ranked BoN on rotating-shift roster. A fired
instance receives ten additional stochastic completions.}
\label{tab:rr_trigger}
\small
\begin{tabular}{@{}lccc@{}}
\toprule
Method & Trigger rate & Mean decodes & Exact feasibility (\%) \\
\midrule
LLaDA-8B (greedy) & -- & 1.00 & 95.3 \\
BoN-10 on every instance & 100\% & 11.00 & 97.8 \\
\rowcolor{xslategray!8}
\textbf{Confidence-triggered Blackboard}
& \textbf{4.7\%} & \textbf{1.47} & \textbf{97.8} \\
\bottomrule
\end{tabular}
\end{table}

Among the $15$ triggered candidate pools, at least one correct roster appears
in $13$ pools, and \textsc{last5} confidence selects a correct roster in $8$.
The trigger therefore recovers more than half of residual greedy errors while
avoiding unconditional sampling.

% ------------------------------------------------------------
% ASYMMETRIC TSP
% ------------------------------------------------------------
\subsection{Asymmetric Traveling Salesperson}
\label{app:atsp_additional}

\subsubsection{Task and Setup}

We additionally evaluate on asymmetric traveling salesperson problems (ATSP),
where the cost from city $i$ to city $j$ need not equal the reverse cost.
Given an $n \times n$ distance matrix, the model outputs an $n \times n$
permutation matrix: row $k$ specifies the city visited in position $k$ of the
tour. The matrix is decoded into a tour beginning and ending at city $0$.

We evaluate $300$ held-out instances, with $100$ each at $n\in\{6,7,8\}$.
Each instance has a unique certified optimum; instances whose best and
second-best tours tie are excluded. We report exact-optimal rate and the rate
within $5\%$ of the optimal tour length. Difficulty is defined independently by
the relative optimality gap,
\[
  \frac{\mathrm{second\_best\_length}-\mathrm{optimal\_length}}
       {\mathrm{optimal\_length}},
\]
where a smaller gap makes the unique optimum harder to distinguish. Instances
are divided into gap quartiles from g0 (largest gap, easiest) to g3 (smallest
gap, hardest).

Blackboard uses the same high-level procedure as JSSP: mean confidence gates
additional computation, but the $K=10$ candidates are ranked by the exact task
objective---minimum valid tour length---rather than by
$\mathcal{C}_\theta$.

\subsubsection{Matched Objective-Selection Results}

Table~\ref{tab:atsp_main} compares matched 8B autoregressive and diffusion
models. Objective-based Best-of-$N$ substantially helps both model families;
the relevant comparison is therefore under the same objective-selection rule.

\begin{table}[h]
\centering
\caption{ATSP results on $300$ held-out instances. Invalid permutation matrices
count as failures for both metrics.}
\label{tab:atsp_main}
\small
\begin{tabular}{@{}lccc@{}}
\toprule
Model and inference & \% optimal & \% within $5\%$ & g3 \% optimal \\
\midrule
LLaMA-3.1-8B (greedy) & 68.7 & 81.3 & 40.8 \\
LLaMA-3.1-8B (objective-BoN-10) & 89.0 & 94.7 & 75.0 \\
\midrule
LLaDA-8B (greedy) & 70.7 & 82.3 & 39.5 \\
\rowcolor{xslategray!8}
\textbf{LLaDA-8B (Blackboard)}
& \textbf{92.7} & \textbf{97.3} & \textbf{82.9} \\
\bottomrule
\end{tabular}
\end{table}

Objective selection improves the autoregressive model from $68.7\%$ to
$89.0\%$ exact optimality. Nevertheless, LLaDA candidates remain higher
quality under the same selection rule, and Blackboard reaches $92.7\%$ exact
optimality.

\subsubsection{Results by Optimality-Gap Difficulty}

\vspace{-1.0em}
\begin{table}[h]
\centering
\caption{LLaDA-8B ATSP exact-optimal rate (\%) by optimality-gap quartile.
g0 has the largest optimality gap; g3 has the smallest gap.}
\label{tab:atsp_gap}
\small
\begin{tabular}{@{}lccccc@{}}
\toprule
Method & All & g0 & g1 & g2 & g3 \\
\midrule
LLaDA-8B (greedy) & 70.7 & 91.9 & 84.0 & 68.0 & 39.5 \\
\rowcolor{xslategray!8}
\textbf{LLaDA-8B (Blackboard)}
& \textbf{92.7} & \textbf{98.6} & \textbf{97.3}
& \textbf{92.0} & \textbf{82.9} \\
\bottomrule
\end{tabular}
\end{table}

The benefit grows as the optimum becomes less distinguishable: Blackboard gains
$+6.7$\,pp on g0 and $+43.4$\,pp on g3. This pattern is consistent with
objective-based candidate selection being most useful when several tours have
similar likelihood but only one is globally optimal.

\subsubsection{Confidence-Triggered Objective BoN}

\vspace{-1.0em}
\begin{table}[h]
\centering
\caption{Compute effect of confidence-triggered objective BoN on ATSP. A fired
instance receives ten additional candidates, selected by minimum tour length.}
\label{tab:atsp_trigger}
\small
\begin{tabular}{@{}lccc@{}}
\toprule
Method & Trigger rate & Mean decodes & \% optimal \\
\midrule
LLaDA-8B (greedy) & -- & 1.0 & 70.7 \\
Objective-BoN-10 on every instance & 100\% & 11.0 & 92.7 \\
\rowcolor{xslategray!8}
\textbf{Confidence-triggered Blackboard}
& \textbf{70\%} & \textbf{8.0} & \textbf{92.7} \\
\bottomrule
\end{tabular}
\end{table}

The trigger fires increasingly often from easy to hard gap bands
($38\%$, $64\%$, $81\%$, and $95\%$ for g0 through g3), and preserves the
exact-optimal rate of unconditional objective-BoN while reducing average
decodes by $27\%$. Here, confidence is used to allocate objective-based search
budget; it is not itself used as the candidate-ranking objective.

\end{document}